\documentclass[runningheads,12pt]{llncs}
\usepackage{helvet}

\usepackage[top=2.5cm, bottom=2.5cm, left=2.4cm, right=2.4cm]{geometry} 
\usepackage{lineno}
\usepackage{amsmath}
\usepackage{amssymb}
\usepackage{graphicx}
\usepackage{booktabs}
\usepackage{listings}
\usepackage{hyperref}
\usepackage{caption}
\usepackage{subcaption}
\usepackage{rotating}
\usepackage{url}
\usepackage{color}
\usepackage[backend=biber,sorting=none,style=numeric-comp]{biblatex}
\usepackage[nomargin,inline,marginclue,draft]{fixme}
\usepackage{cleveref}
\usepackage[font=small]{caption}
\usepackage[labelformat=empty, position=top]{subcaption}
\usepackage[export]{adjustbox}
\usepackage{multirow}
\usepackage[table,xcdraw]{xcolor}
\fxsetup{status=draft}
\fxsetup{theme=color, mode=multiuser} \FXRegisterAuthor{me}{ame}{\color{red}Me}
\newcommand{\orcid}[1]{\href{https://orcid.org/#1}{\includegraphics[scale=0.02]{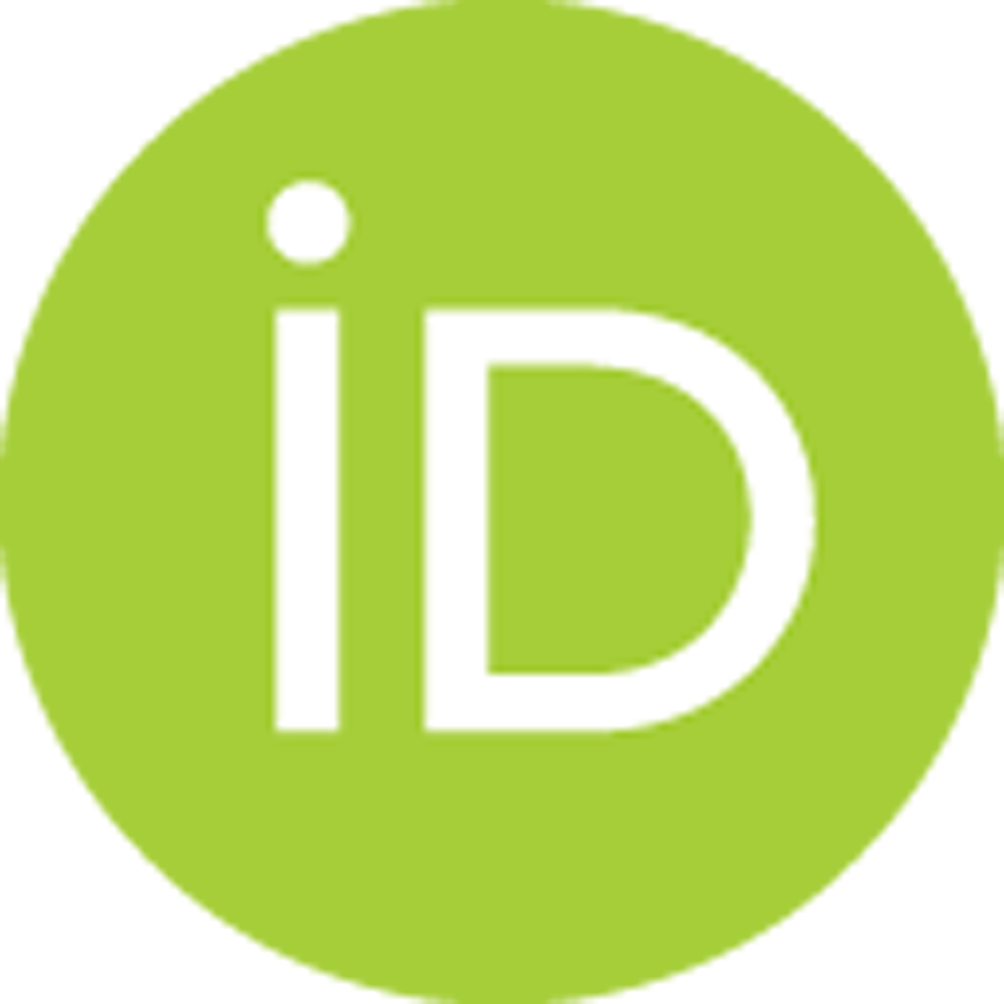}}} 
\hypersetup{
    colorlinks=true,
    citecolor=blue,
    urlcolor=purple,
}
\let\cite=\supercite

\title{Federated Learning Enables Big Data for Rare Cancer Boundary Detection}
\titlerunning{FL Enabling Big Data}

\author{
\footnotesize{Sarthak Pati\inst{1,2,3,4,*,\orcid{0000-0003-2243-8487}}
\and
Ujjwal Baid\inst{1,2,3,*,\orcid{0000-0001-5246-2088}}
\and
Brandon Edwards\inst{5,*,\orcid{0000-0002-0957-9149}}
\and
Micah Sheller\inst{5,\orcid{0000-0002-6571-0850}}
\and
Shih-Han Wang\inst{5,\orcid{0000-0001-9713-3878}}
\and
G Anthony Reina\inst{5,\orcid{0000-0001-9623-9259}}
\and
Patrick Foley\inst{5,\orcid{0000-0001-9401-3088}}
\and
Alexey Gruzdev\inst{5,\orcid{0000-0003-3924-4531}}
\and
Deepthi Karkada\inst{5,\orcid{0000-0002-0623-548X}}
\and
Christos Davatzikos\inst{1,2,\orcid{0000-0002-1025-8561}}
% ...
}
}

\authorrunning{Pati et al.}

\institute{
% ...
}

\begin{document}
\sloppy 
\mainmatter
\maketitle
\setcounter{footnote}{0}
\newpage


\begin{abstract}

    \textbf{Although machine learning (ML) has shown promise in numerous domains, there are concerns about generalizability to out-of-sample data. This is currently addressed by centrally sharing ample, and importantly diverse, data from multiple sites. However, such centralization is challenging to scale (or even not feasible) due to various limitations. Federated ML (FL) provides an alternative to train accurate and generalizable ML models, by only sharing numerical model updates. Here we present findings from the largest FL study to-date, involving data from $71$ healthcare institutions across $6$ continents, to generate an automatic tumor boundary detector for the rare disease of glioblastoma, utilizing the largest dataset of such patients ever used in the literature ($25,256$ MRI scans from $6,314$ patients). We demonstrate a $33\%$ improvement over a publicly trained model to delineate the surgically targetable tumor, and $23\%$ improvement over the tumor's entire extent. We anticipate our study to: 1) enable more studies in healthcare informed by large and diverse data, ensuring meaningful results for rare diseases and underrepresented populations, 2) facilitate further quantitative analyses for glioblastoma via performance optimization of our consensus model for eventual public release, and 3) demonstrate the effectiveness of FL at such scale and task complexity as a paradigm shift for multi-site collaborations, alleviating the need for data sharing.}

 \end{abstract}



    Advances in machine learning (\textbf{\textit{ML}}), and particularly deep learning (\textbf{\textit{DL}}), have shown promise in addressing complex healthcare problems \cite{akbari2014pattern,akbari2016imaging,bakas2017vivo,akbari2018vivo,binder2018epidermal,rathore2018deriving,bakas2018nimg,pisapia2018use,davatzikos2019precision,fathi2020imaging,bakas2020overall,akbari2020histopathology,mang2020integrated,menze2021analyzing}. However, there are concerns about generalizability on data from sources that did not participate in model training, i.e., ``out-of-sample'' data \cite{maartensson2020reliability,zech2018variable}. Literature indicates that training robust and accurate models requires large amounts of data \cite{obermeyer2016predicting,marcus2018deep,aggarwal2018neural}, the diversity of which affects model generalizability to ``out-of-sample'' cases \cite{thompson2014enigma}. To address these concerns, models need to be trained on data originating from numerous sites representing diverse population samples. The current paradigm for such multi-site collaborations is ``centralized learning'' (\textbf{\textit{CL}}), in which data from different sites are shared to a centralized location following inter-site agreements \cite{thompson2014enigma,glass2018glioma,davatzikos2020ai,bakas2020iglass}. However, such data centralization is difficult to scale (and might not even be feasible), especially at a global scale, due to concerns \cite{rieke2020future,sheller2020federated} relating to privacy, data-ownership, intellectual property, technical challenges (e.g., network and storage limitations), as well as compliance with varying regulatory policies (e.g., Health Insurance Portability and Accountability Act (HIPAA) of the United States \cite{annas2003hipaa} and the General Data Protection Regulation (GDPR) of the European Union \cite{voigt2017eu}). In contrast to this centralized paradigm, ``federated learning'' (\textbf{\textit{FL}}) describes an approach where models are trained by only sharing model parameter updates from decentralized data (i.e., each site retains its data locally) \cite{mcmahan2017communication,sheller2018multiinstitutional,sheller2020federated,rieke2020future,dayan2021federated}, without sacrificing performance when compared to CL-trained models \cite{chang2018distributed,nilsson2018performance,sheller2018multiinstitutional,sheller2020federated,sarma2021federated,shen2021multi,yang2021federated}. Thus, FL can offer an alternative to CL, potentially creating a paradigm shift that alleviates the need for data sharing, and hence increase access to geographically-distinct collaborators, thereby increasing the size and diversity of data used to train ML models.

\begin{figure}
    \begin{subfigure}[t]{0.70\textwidth}
        \begin{subfigure}[t]{0.03\textwidth}
        \textbf{a}
        \end{subfigure}
        \begin{subfigure}[t]{0.96\textwidth}
            \includegraphics[width=\textwidth, valign=t]{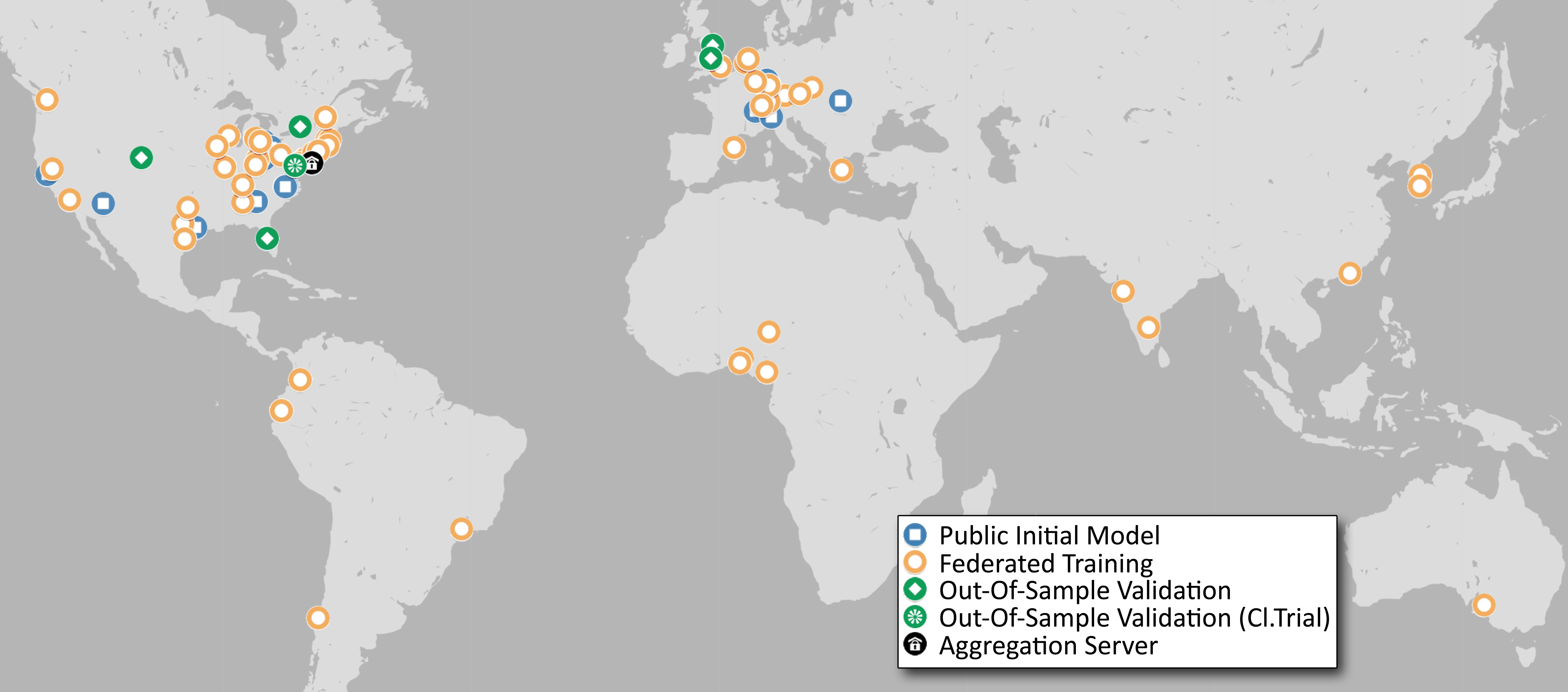}
        \end{subfigure}
        \\
        \begin{subfigure}[t]{0.03\textwidth}
        \textbf{b}
        \end{subfigure}
        \begin{subfigure}[t]{0.25\textwidth}
            \centering
            \includegraphics[height=11cm, valign=t]{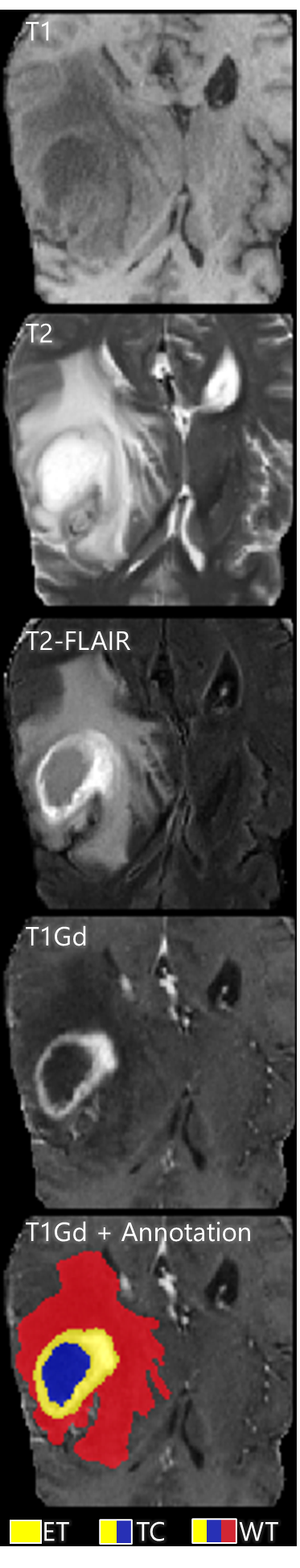}
        \end{subfigure}
        \hfill
        \begin{subfigure}[t]{0.70\textwidth}
            \begin{subfigure}[t]{0.03\textwidth}
            \textbf{c}
            \end{subfigure}
                \begin{subfigure}[t]{0.96\textwidth}
                    \includegraphics[width=\textwidth, valign=t]{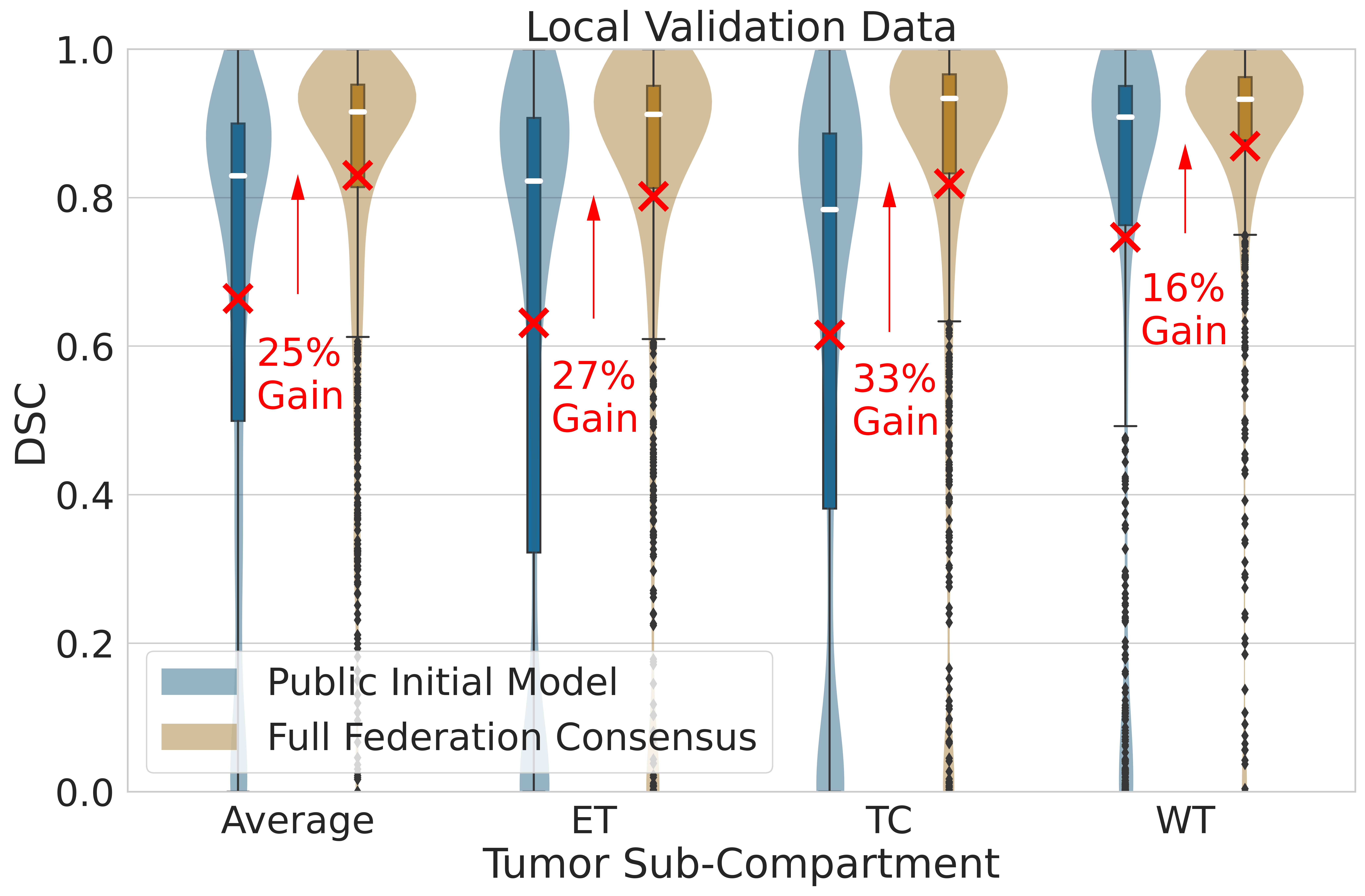}
                \end{subfigure}
            \\
            \begin{subfigure}[t]{0.03\textwidth}
            \textbf{d}
            \end{subfigure}
                \begin{subfigure}[t]{0.94\textwidth}
                    \includegraphics[width=\textwidth, valign=t]{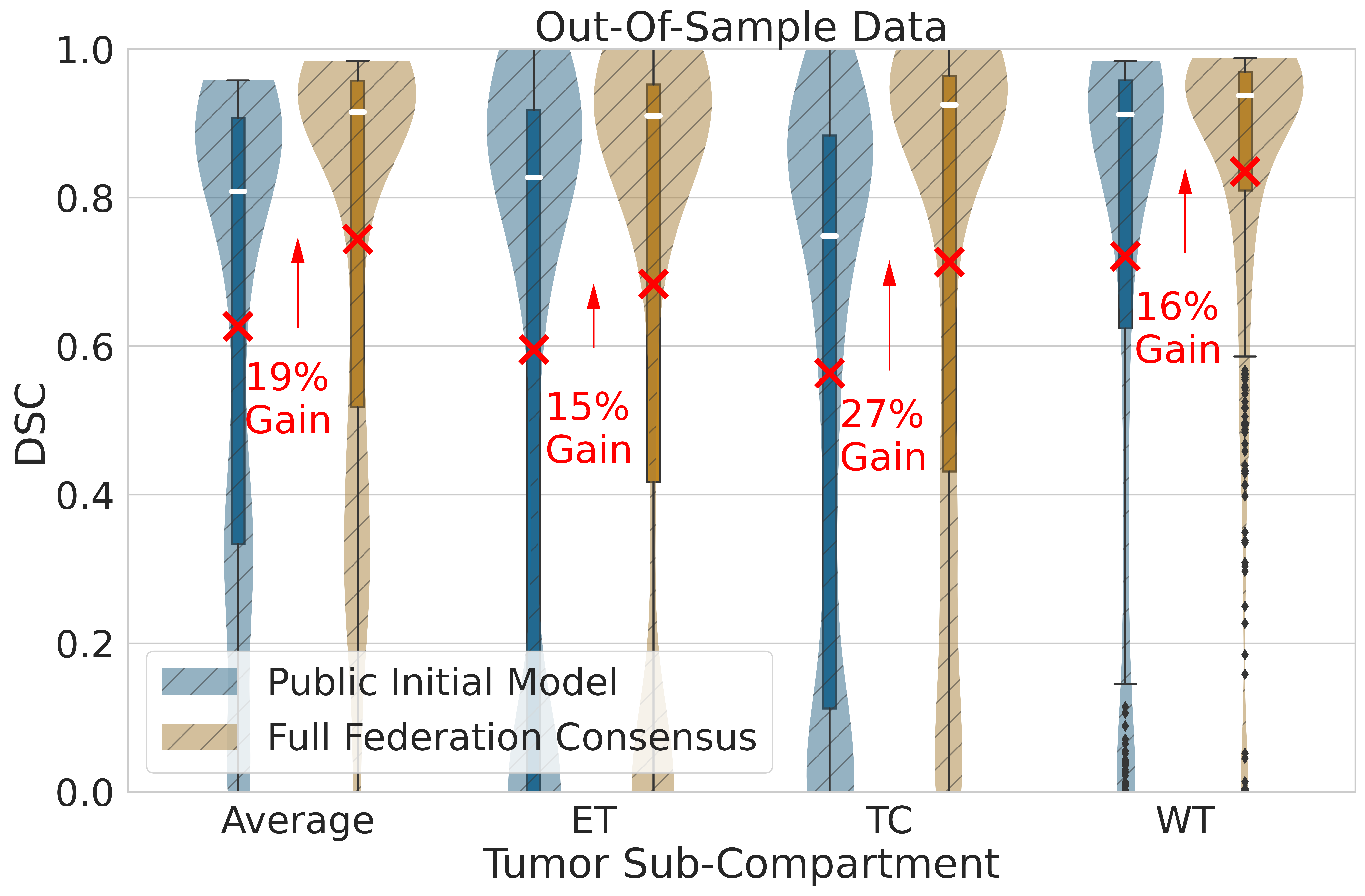}
                \end{subfigure}
        \end{subfigure}
    \end{subfigure}
    \hfill
    \begin{subfigure}[t]{0.03\textwidth}
    \textbf{e}
    \end{subfigure}
    \begin{subfigure}[t]{0.24\textwidth}
        \centering
        \includegraphics[width=\textwidth, valign=t]{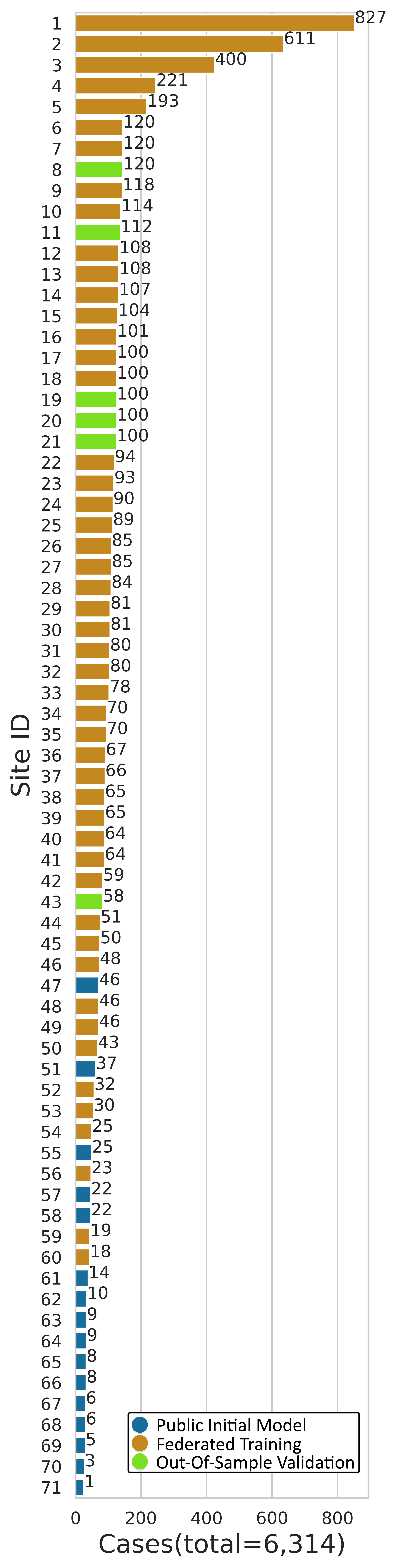}
    \end{subfigure}
    \caption{Representation of the study's global scale, diversity, and complexity. \textbf{a}, The map of all sites involved in the development of FL consensus model. \textbf{b}, example of a glioblastoma mpMRI scan with corresponding reference annotations of the tumor sub-compartments. \textbf{c-d}, comparative performance evaluation of the final consensus model with the public initial model on the collaborators' local validation data (in \textbf{c}) and on the complete out-of-sample data (in \textbf{d}), per tumor sub-compartment. Note the box and whiskers inside each violin plot, represent the true min and max values. The top and bottom of each ``box'' depict the 3rd and 1st quartile of each measure. The white line and the red $\times$, within each box, indicate the median and mean values, respectively. The fact that these are not necessarily at the centre of each box indicates the skewness of the distribution over different cases. The ``whiskers'' drawn above and below each box depict the extremal observations still within $1.5$ times the interquartile range, above the 3rd or below the 1st quartile. \textbf{e}, number of contributed cases per collaborating site.}
    \label{fig:map}
    \label{fig:benefits_of_fed}
    \label{fig:benefits_of_fed_against_out_of_sample} 
    \label{fig:mpmri_and_labels} 
\end{figure}

FL has tremendous potential in healthcare \cite{natMed2018Retina,natMed2019Cardiac}, particularly towards addressing health disparities, under-served populations, and ``rare'' diseases \cite{griggs2009rareDisease}, by enabling ML models to gain knowledge from ample and diverse data that would otherwise not be available. With that in mind, here we focus on the ``rare'' disease of glioblastoma, and particularly on the detection of its extent using multi-parametric magnetic resonance imaging (mpMRI) scans \cite{shukla2017advanced}. While glioblastoma is the most common malignant primary brain tumor \cite{brennan2013somatic,verhaak2010integrated,sottoriva2013intratumor}, it is still classified as a ``rare'' disease, as its incidence rate (i.e., $~3$/$100,000$ people) is substantially lower than the rare disease definition rate (i.e., $<10$/$100,000$ people) \cite{griggs2009rareDisease}. This means that single sites cannot collect large and diverse datasets to train robust and generalizable ML models, and necessitates collaboration between geographically distinct sites. Despite extensive efforts to improve prognosis of glioblastoma patients with intense multimodal therapy, their median overall survival is only $14.6$ months after standard-of-care treatment, and $4$ months without treatment \cite{ostrom2019cbtrus}. Although the subtyping of glioblastoma has been improved \cite{louis20212021} and the standard-of-care treatment options have expanded during the last twenty years, there have been no substantial improvements in overall survival \cite{han2020deep}. This reflects the major obstacle in treating these tumors that is their intrinsic heterogeneity \cite{brennan2013somatic,sottoriva2013intratumor}, and the need for analyses of larger and more diverse data towards better understanding the disease. In terms of radiologic appearance, glioblastomas comprise $3$ main sub-compartments, defined as i) the ``enhancing tumor'' (ET), representing the vascular blood-brain barrier breakdown within the tumor, ii) the ``tumor core'' (TC), which includes the ET and the necrotic (NCR) part, and represents the surgically relevant part of the tumor, and iii) the ``whole tumor'' (WT), which is defined by the union of the TC and the peritumoral edematous/infiltrated tissue (ED), and represents the complete tumor extent relevant to radiotherapy (Fig.~\ref{fig:mpmri_and_labels}.b). Detecting these sub-compartment boundaries therefore defines a multi-parametric multi-class learning problem \cite{bakas2015glistrboost,bakas2015segmentation,zeng2016segmentation,rudie2019multi,pei2020longitudinal}, and is a critical first step towards further quantifying and assessing this heterogeneous rare disease and ultimately influencing clinical decision-making.

Co-authors in this study have previously introduced FL in healthcare in a simulated setting \cite{sheller2018multiinstitutional} and evaluated different FL approaches \cite{sheller2020federated} for the same use case as the present study, i.e., detecting the boundaries of glioblastoma sub-compartments. Findings from these studies supported the superiority of the FL approach used in the present study (i.e., based on an aggregation server \cite{mcmahan2017communication,rieke2020future}), which had almost identical performance to CL. The present study describes the largest to-date global FL effort to develop an accurate and generalizable ML model for detecting glioblastoma sub-compartment boundaries, based on $25,256$ MRI scans (over 5TB) of $6,314$ glioblastoma patients from $71$ geographically distinct sites, across $6$ continents (Fig.~\ref{fig:map}.a). Notably, this describes the largest and most diverse dataset of glioblastoma patients ever considered in the literature. It was the use of FL that successfully enabled our ML model to gain knowledge from such an unprecedented dataset. The extended global footprint and the task complexity is what sets this study apart from current literature, since it dealt with a multi-parametric multi-class problem with reference standards that require expert clinicians following an involved manual annotation protocol, rather than recording a categorical entry from medical records \cite{roth2020federated,dayan2021federated}. Moreover, varying characteristics of the mpMRI data due to scanner hardware and acquisition protocol differences \cite{chaichana2013multi,fathi2020cancer} were handled at each collaborating site via established harmonized preprocessing pipelines \cite{brats_1,brats_2,brats_3,brats_4}.

The scientific contributions of this manuscript can be summarized by i) the insights garnered during this work that can pave the way for more successful FL studies of increased scale and task complexity, ii) making a potential impact for the treatment of the rare disease of glioblastoma, by eventually publicly releasing an optimized trained consensus model for use in resource-constrained clinical settings, and iii) demonstrating the effectiveness of FL at such scale and task complexity as a paradigm shift redefining multi-site collaborations, while alleviating the need for data sharing.



\section*{Methods}

\subsection*{Data}
\label{methods:data}

    The data considered in this study described patient populations with adult-type diffuse glioma \cite{louis20212021}, and specifically displaying the radiological features of glioblastoma, scanned with mpMRI to characterize the anatomical tissue structure \cite{shukla2017advanced}. Each case is specifically described by i) native T1-weighted (T1), ii) Gadolinium-enhanced T1-weighted (T1Gd), iii) T2-weighted (T2), and iv) T2-weighted-Fluid-Attenuated-Inversion-Recovery (T2-FLAIR) MRI scans. Cases with any of these sequences missing were not included in the study. Note that no inclusion/exclusion criterion applied relating to the type of acquisition (i.e., both 2D axial and 3D acquisitions were included, with a preference for 3D if available), or the exact type of sequence (e.g., MP-RAGE vs SPGR). The only exclusion criterion was for T1-FLAIR scans that were intentionally excluded to avoid mixing varying tissue appearance due to the type of sequence, across native T1-weighted scans.

    The eligibility of collaborating sites to participate in the federation was determined based on data availability, and approval by their respective institutional review board. $55$ sites participated as independent collaborators in the study defining a dataset of $6,083$ cases. The MRI scanners used for data acquisition were from multiple vendors (i.e., Siemens, GE, Philips, Hitachi, Toshiba), with magnetic field strength ranging from 1T to 3T. The data from all $55$ collaborating sites followed a male:female ratio of $1.47:1$ with age ranging between $7$ and $94$ years. 
    
    From all $55$ collaborating sites, $49$ were chosen to be part of the training phase, and $6$ sites were categorized as ``out-of-sample'', i.e., none of these were part of the training stage. These specific $6$ out-of-sample sites (Site IDs: 8, 11, 19, 20, 21, 43) were allocated based on their availability, i.e., they have indicated an expected delayed participation rendering them optimal for model generalizability validation. One of these $6$ out-of-sample sites (Site 11) contributed aggregated \textit{a priori} data from a multi-site randomized clinical trial for newly diagnosed glioblastoma (ClinicalTrials.gov Identifier: NCT00884741, RTOG0825 \cite{RTOG_2013_abstract,RTOG0825_2014_NEJM}, ACRIN6686 \cite{ACRIN6686_2018_NOnc,ACRIN6686_2021_NOnc}), with inherent diversity benefiting the intended generalizability validation purpose. The American College of Radiology (ACR - Site 11) serves as the custodian of this trial's imaging data on behalf of ECOG-ACRIN, who made the data available for this study. Following screening for availability of the $4$ required mpMRI scans with sufficient signal-to-noise ratio judged by visual observation, a subset of $362$ cases from the original trial data were included in this study. The out-of-sample data totaled $590$ cases intentionally held-out of the federation, with the intention of validating the consensus model in completely unseen cases. To facilitate further such generalizability evaluation without burdening the collaborating sites, a subset consisting of $332$ cases (including the multi-site clinical data provided by ACR) from this out-of-sample data was aggregated, to serve as the ``\textit{centralized out-of-sample}'' dataset. Furthermore, the $49$ sites participating in the training phase define a collective dataset of $5,493$ cases. The exact $49$ site IDs are: 1, 2, 3, 4, 5, 6, 7, 9, 10, 12, 13, 14, 15, 16, 17, 18, 22, 23, 24, 25, 26, 27, 28, 29, 30, 31, 32, 33, 34, 35, 36, 37, 38, 39, 40, 41, 42, 44, 45, 46, 48, 49, 50, 52, 53, 54, 56, 59, 60. These cases were automatically split at each site following a $4:1$ ratio between cases for training and local validation. During the federated training phase the data used for the public initial model were also included as a dataset from a separate node, such that the contribution of sites providing the publicly available data is not forgotten within the global consensus model. This results in the final consensus model being developed based on data from $71$ sites over a total dataset of $6,314$ cases.

\subsection*{Harmonized Data Preprocessing}
\label{methods:dataPreProc}

    Once each collaborating site identified their local data, they were asked to use the preprocessing functionality of the software platform we provided. This functionality follows the harmonized data preprocessing protocol defined by the BraTS challenge \cite{brats_1,brats_2,brats_3,brats_4}, 
    to account for inter-site acquisition protocol variations, e.g., 3D vs 2D axial plane acquisitions.

\subsection*{The neural network architecture}
\label{methods:network}

    The trained model to delineate the different tumor sub-compartments was based on the popular 3D U-Net with residual connections (3D-ResUNet) \cite{ronneberger2015u,cciccek20163d,he2016deep,drozdzal2016importance,bhalerao2019brain}. The network had $30$ base filters, with a learning rate of $lr=5\times10^{-5}$ optimized using the Adam optimizer \cite{kingma2014adam}. For the loss function used in training, we used the generalized $DSC$ score \cite{sudre2017generalised,dice} (represented mathematically in Eq.~\ref{eq:dice}) on the absolute complement of each tumor sub-compartment independently. Such mirrored $DSC$ loss has been shown to capture variations in smaller regions better \cite{isensee2021nnu}. No penalties were used in the loss function, due to our use of `mirrored' $DSC$ loss \cite{chen2016efficient,salehi2017tversky,caliva2019distance}. The final layer of the model was a sigmoid layer, providing three channel outputs for each voxel in the input volume, one output channel per tumor sub-compartment. While the generalized $DSC$ score was calculated using a binarized version of the output (check sigmoid value against the threshold $0.5$) for the final prediction, we used the floating point $DSC$ \cite{shamir2019continuous} during the training process.
    
    \begin{equation}
        \label{eq:dice}
        DSC = \frac{2\left\|RL \odot PM \right\|_1}{\left\|RL \right\|_1 + \left\|PM \right\|_1}
    \end{equation}
    \noindent where RL serves as the reference label, $PM$ is the predicted mask, $\odot$ is the Hadamard product \cite{horn1990hadamard} (i.e., component wise multiplication), and $\left\| x \right\|_1$ is the L1-norm \cite{barrodale1968l1}, i.e., sum of the absolute values of all components).
    
\subsection*{The Federation}
\label{methods:federation}

    The collaborative network of the present study spans $6$ continents (Fig.~\ref{fig:map}), with data from $71$ geographically distinct sites. The training process was initiated when each collaborator securely connected to a central aggregation server, which resided behind a firewall at the University of Pennsylvania. As soon as the secure connection was established, the public initial model was passed to the collaborating site. Using FL based on an aggregation server, collaborating sites then trained the same network architecture on their local data for one epoch, and shared model updates with the central aggregation server. The central aggregation server received model updates from all collaborators, combined them (by averaging model parameters) and sent the consensus model back to each collaborator to continue their local training. Each such iteration is called a ``\textit{federated round}''. After not observing any meaningful changes since round $42$, we stopped the training after a total of $73$ federated rounds. Additionally, we performed all operations on the aggregator on secure hardware (by leveraging trusted execution environments \cite{kaissis2020secure}), in order to increase the trust by all parties in the confidentiality of the model updates being computed and shared, as well as to increase the confidence in the integrity of the computations being performed \cite{ekberg2014untapped}.
    
    The federated training was initialized using a ``\textit{public initial model}'' trained on $231$ cases from $16$ sites. This was done as opposed to a random initialization point to facilitate faster convergence of the model performance \cite{raghu2019transfusion,young2020applicability}. After the training process was complete, the ``\textit{final consensus model}'' was obtained after model selection from all the global consensus models obtained for each federated round. Thus, the final consensus model was developed on $6,314$ cases from $71$ sites. To quantitatively evaluate the performance of the trained models, $20\%$ of the total cases contributed by each participating site were excluded from the model training process and were used as ``\textit{local validation data}''. To further evaluate the generalizability of the models in unseen data, $6$ sites were not involved in any of the training stages to represent an unseen ``out-of-sample'' data population of $590$ cases. To facilitate further evaluation without burdening the collaborating sites, a subset ($n=332$) of these cases was aggregated to serve as a ``\textit{centralized} out-of-sample'' dataset. Model performance was quantitatively evaluated here using the Dice Similarity Coefficient ($DSC$), which assesses the spatial agreement between the model's prediction and the reference standard for each of the $3$ tumor sub-compartments (ET, TC, WT).

\subsection*{Model Run-time in 
Clinical Environments}
\label{methods:optimization}

    Clinical environments typically have constrained computational resources, such as the availability of specialized hardware (e.g., DL acceleration cards) and increased memory, which affect the run-time performance of DL inference workloads. Thus, taking into consideration the potential deployment of the final consensus model in such low-resource settings, we decided to proceed with a single 3D-ResUNet, rather than an ensemble of multiple models. This decision ensured a reduced computational burden, when compared with running multiple models, which is typically done in academic research projects \cite{brats_1,brats_2,brats_3,brats_4}.
    
    To further facilitate use in low-resource environments, we plan to publicly release run-time optimized \cite{rodriguez2018lower} version of the final consensus model. For this model, graph level optimizations (i.e., operators fusion) were initially applied, followed by optimizations for low precision inference, i.e., converting the floating point single precision model to a fixed precision 8-bit integer model (a process known as ``quantization'' \cite{lin2016fixed}). In particular, we used accuracy-aware quantization \cite{vakili2013enhanced}, where model layers were iteratively scaled to a lower precision format. These optimizations yielded several run-time performance benefits, such as lower inference latency (a platform-dependent $4.48\times$ average speedup and $2.29\times$ reduced memory requirement when compared with the original consensus model), and higher throughput (equal to the $4.48\times$ speedup improvement since the batch size used is equal to $1$), while the trade-off was an insignificant ($p_{Average}<7\times10^{-5}$) drop in the average $DSC$.

\subsection*{Code availability}

    Motivated by findability, accessibility, interoperability, and reusability (FAIR) criteria in scientific research \cite{wilkinson2016fair}, all the code used to design the Federated Tumor Segmentation (FeTS) platform for this study is available at \url{https://github.com/FETS-AI/Front-End}. The functionality related to preprocessing (i.e., DICOM to NIfTI conversion, population-based harmonized preprocessing, co-registration) and manual refinements of annotation is derived from the open-source Cancer Imaging Phenomics Toolkit (CaPTk, \url{https://github.com/CBICA/CaPTk}) \cite{captk_1,captk_2,captk_3}. The co-registration is performed using the Greedy framework \cite{yushkevich2016fast}, available via CaPTk \cite{captk_1,captk_2,captk_3}, ITK-SNAP \cite{py06nimg}, and the FeTS tool. The brain extraction \cite{han2018brain} is done using the BrainMaGe method \cite{thakur2020brain}, and is available in \url{https://github.com/CBICA/BrainMaGe} and via the Generally Nuanced Deep Learning Framework (GaNDLF) \cite{pati2021gandlf} at \url{https://github.com/CBICA/GaNDLF}. To generate automated annotations, DeepMedic's \cite{kamnitsas2017efficient} integration with CaPTk was used, and we used the model weights and inference mechanism provided by the other algorithm developers (DeepScan \cite{mckinley2018ensembles} and nnU-Net \cite{isensee2021nnu} (\url{https://github.com/MIC-DKFZ/nnunet})). DeepMedic's original implementation is available in \url{https://github.com/deepmedic/deepmedic}, whereas the one we used in this study can be found at \url{https://github.com/CBICA/deepmedic}. The fusion of the labels was done using the Label Fusion tool \cite{labelfusion} available at \url{https://github.com/FETS-AI/LabelFusion}. The data loading pipeline and network architecture was developed using the GaNDLF framework \cite{pati2021gandlf} by using PyTorch \cite{paszke2019pytorch}. The data augmentation was done via GaNDLF by leveraging TorchIO \cite{garcia_torchio_2020}. The FL backend developed for this project has been open-sourced as a separate software library, to encourage further research on FL \cite{reina2021openfl} and is available at \url{https://github.com/intel/openfl}. The optimization of the consensus model inference workload was performed via OpenVINO \cite{gorbachev2019openvino} (\url{https://github.com/openvinotoolkit/openvino/tree/2021.4.1}), which is an open-source toolkit enabling acceleration of neural network models through various optimization techniques. The optimizations were evaluated on an Intel Core\textsuperscript{\textregistered} i7-1185G7E CPU @ 2.80GHz with $2 \times 8$ GB DDR4 3200MHz memory on Ubuntu 18.04.6 OS and Linux kernel version 5.9.0-050900-generic.

\section*{Results}

    At the time of initialization of the federation, the public initial model was evaluated against the local validation data of all sites, resulting in an average (across all cases of all sites) $DSC$ per sub-compartment, of: $DSC_{ET}=0.63$, $DSC_{TC}=0.62$, $DSC_{WT}=0.75$. To summarize the model performance with a single collective score, we then calculate the average $DSC$ (across all $3$ tumor sub-compartments per case, and then across all cases of all sites) as equal to $0.66$. 
    
    Following model training across all sites, the final consensus model garnered significant performance improvements against the collaborators' local validation data of $27\%$ ($p_{ET}<1\times10^{-36}$), $33\%$ ($p_{TC}<1\times10^{-59}$), and $16\%$ ($p_{WT}<1\times10^{-21}$), for ET, TC, and WT, respectively (Fig.~\ref{fig:benefits_of_fed}.c). 
    
    To further evaluate the potential generalizability improvements of the final consensus model on unseen data, we compared it with the public initial model against the complete out-of-sample data, and noted significant performance improvements of $15\%$ ($p_{ET}<1\times10^{-5}$), $27$\% ($p_{TC}<1\times10^{-16}$), and $16\%$ ($p_{WT}<1\times10^{-7}$), for ET, TC, and WT, respectively (Fig.~\ref{fig:benefits_of_fed_against_out_of_sample}.d). 
    
    Notably, the only difference between the public initial model and the final consensus model, was that the latter gained knowledge during training from increased datasets contributed by the complete set of collaborators. The conclusion of these findings reinforces the importance of using large and diverse data for generalizable models to ultimately drive patient care.



\section*{Discussion}
In this study, we have described the largest real-world FL effort to-date utilizing data of $6,314$ glioblastoma patients from $71$ geographically unique sites spread across $6$ continents, to develop an accurate and generalizable ML model for detecting glioblastoma sub-compartment boundaries. Notably, this extensive global footprint of the collaborating sites in this study, also yield the largest dataset ever reported in the literature assessing this rare disease. It is the use of FL that successfully enabled \textit{i)} access to such an unprecedented dataset of the most common and fatal adult brain tumor, and \textit{ii)} meaningful ML training to ensure generalizability of models across out-of-sample data. Since glioblastoma boundary detection is critical for treatment planning and the requisite first step for further quantitative analyses, the models generated during this study have the potential to make far-reaching clinical impact.

The large and diverse data that FL enabled, led to the final consensus model garnering significant performance improvements over the public initial model against both the collaborators' local validation data and the complete out-of-sample data. The improved result is a clear indication of the benefit that can be afforded through access to more data. In comparison with the limited existing real-world FL studies \cite{roth2020federated,dayan2021federated}, our use case is larger in scale and substantially more complex, since it 1) addresses a multi-parametric multi-class problem, with reference standards that require expert collaborating clinicians to follow an involved manual annotation protocol, rather than simply recording a categorical entry from medical records, and 2) requires the data to be preprocessed in a harmonized manner to account for differences in MRI acquisition. 

We have demonstrated the utility of an FL approach to develop an accurate and generalizable ML model for detecting glioblastoma sub-compartment boundaries, a finding which is of particular relevance for neurosurgical and radiotherapy planning in patients with this disease. This study is meant to be used as a blueprint for future FL studies that result in clinically deployable ML models. Building on this study, a continuous FL consortium would enable downstream quantitative analyses with implications for both routine practice and clinical trials, and most importantly, increase access to high-quality precision care worldwide. Furthermore, the lessons learned from this study with such a global footprint are invaluable and can be applied to a broad array of clinical scenarios with the potential for great impact to rare diseases and underrepresented populations.











\section*{Acknowledgments}
    Research and main methodological developments reported in this publication were partly supported by the National Institutes of Health (NIH) under award numbers NIH/NCI:U01CA242871 (S.Bakas), NIH/NINDS:R01NS042645 (C.Davatzikos), NIH/NCI:U24CA189523 (C.Davatzikos), NIH/NCI:U24CA215109 (J.Saltz), NIH/NCI:U01CA248226 (P.Tiwari), NIH/NCI:P30CA51008 (Y.Gusev), NIH:R50CA211270 (M.Muzi), NIH/NCATS:UL1TR001433 (Y.Yuan), NIH/NIBIB:R21EB030209 (Y.Yuan), NIH/NCI:R37CA214955 (A.Rao), and NIH:R01CA233888 (A.L.Simpson). The authors would also like to acknowledge the following NIH funded awards for the multi-site clinical trial (NCT00884741, RTOG0825/ACRIN6686): U10CA21661, U10CA37422, U10CA180820, U10CA180794, U01CA176110, R01CA082500, CA079778, CA080098, CA180794, CA180820, CA180822, CA180868. Research reported in this publication was also partly supported by the National Science Foundation, under award numbers 2040532 (S.Baek), and 2040462 (B.Landman). Research reported in this publication was also supported by 
    i) a research grant from Varian Medical Systems (Palo Alto, CA USA) (Y.Yuan),
    ii) the Ministry of Health of the Czech Republic (Grant Nr. NU21-08-00359) (M.Ke\v{r}kovsk\'{y} and M.Kozubek),
    iii) Deutsche Forschungsgemeinschaft (DFG, German Research Foundation) Project-ID 404521405, SFB 1389, Work Package C02, and Priority Programme 2177 “Radiomics: Next Generation of Biomedical Imaging” (KI 2410/1-1 | MA 6340/18-1) (P.Vollmuth),
    iv) DFG Project-ID B12, SFB 824 (B.Wiestler),
    v) the Helmholtz Association (funding number ZT-I-OO1 4) (K.Maier-Hein),
    vi) the Dutch Cancer Society (KWF project number EMCR 2015-7859) (S.R.van der Voort),
    vii) the Chilean National Agency for Research and Development (ANID-Basal FB0008 (AC3E) and FB210017 (CENIA)) (P.Guevara),
    viii) the Canada CIFAR AI Chairs Program (M.Valli\`{e}res),
    ix) Leeds Hospital Charity (Ref: 9RO1/1403) (S.Currie),
    x) the Cancer Research UK funding for the Leeds Radiotherapy Research Centre of Excellence (RadNet) and the grant number C19942/A28832 (S.Currie),
    xi) Medical Research Council (MRC) Doctoral Training Programme in Precision Medicine (Award Reference No. 2096671) (J.Bernal),
    xii) The European Research Council (ERC) under the European Union’s Horizon 2020 research and innovation programme (Grant Agreement No. 757173) (B.Glocker),
    xiii) The UKRI London Medical Imaging \& Artificial Intelligence Centre for Value Based Healthcare (K.Kamnitsas),
    xiv) Wellcome/Engineering and Physical Sciences Research Council (EPSRC) Center for Medical Engineering (WT 203148/Z/16/Z) (T.C.Booth), 
    xv) American Cancer Society Research Scholar Grant RSG-16-005-01 (A.Rao),
    xvi) the Department of Defense (DOD) Peer Reviewed Cancer Research Program (PRCRP) W81XWH-18-1-0404, Dana Foundation David Mahoney Neuroimaging Program, the V Foundation Translational Research Award, Johnson \& Johnson WiSTEM2D Award (P.Tiwari),
    xvii) RSNA Research \& Education Foundation under grant number RR2011 (E.Calabrese),
    xviii) the National Research Fund of Luxembourg (FNR) (grant number: C20/BM/14646004/GLASS-LUX/Niclou) (S.P.Niclou),
    xix) EU Marie Curie FP7-PEOPLE-2012-ITN project TRANSACT (PITN-GA-2012-316679) and the Swiss National Science Foundation (project number 140958) (J.Slotboom), and
    xx) CNPq 303808/2018-7 and FAPESP 2014/12236-1 (A.Xavier Falc\~{a}o). The content of this publication is solely the responsibility of the authors and does not represent the official views of the NIH, the NSF, the RSNA R\&E Foundation, or any of the additional funding bodies.
    
    
\section*{Author Contributions}
    
    \noindent\underline{\textbf{Study Conception:}} S.Pati, U.Baid, B.Edwards, M.Sheller, G.A.Reina, J.Martin, S.Bakas.
    
    \noindent \underline{\textbf{Development of software used in the study:}} S.Pati, B.Edwards, M.Sheller, S.Wang, G.A.Reina, P.Foley, A.Gruzdev, D.Karkada.
    
    \noindent \underline{\textbf{Data Acquisition:}} M.Bilello, S.Mohan, E.Calabrese, J.Rudie, J.Saini, R.Y.Huang, K.Chang, T.So, P.Heng, T.F.Cloughesy, C.Raymond, T.Oughourlian, A.Hagiwara, C.Wang, M.To, M.Ke\v{r}kovsk\'{y}, T.Kop\v{r}ivov\'{a}, M.Dost\'{a}l, V.Vyb\'{i}hal, J.A.Maldjian, M.C.Pinho, D.Reddy, J.Holcomb, B.Wiestler, M.Metz, R.Jain, M.Lee, P.Tiwari, R.Verma, Y.Gusev, K.Bhuvaneshwar, C.Bencheqroun, A.Belouali, A.Abayazeed, A.Abbassy, S.Gamal, M.Qayati, M.Mekhaimar, M.Reyes, R.R.Colen, M.Ak, P.Vollmuth, G.Brugnara, F.Sahm, M.Bendszus, W.Wick, A.Mahajan, C.Bala\~{n}a Quintero, J.Capellades, J.Puig, Y.Choi, M.Muzi, H.F.Shaykh, A.Herrera-Trujillo, W.Escobar, A.Abello, P.LaMontagne, B.Landman, K.Ramadass, K.Xu, S.Chotai, L.B.Chambless, A.Mistry, R.C.Thompson, J.Bapuraj, N.Wang, S.R.van der Voort, F.Incekara, M.M.J.Wijnenga, R.Gahrmann, J.W.Schouten, H.J.Dubbink, A.J.P.E.Vincent, M.J.van den Bent, H.I.Sair, C.K.Jones, A.Venkataraman, J.Garrett, M.Larson, B.Menze, T.Weiss, M.Weller, A.Bink, B.Pouymayou, Y.Yuan, S.Sharma, T.Tseng, B.C.A.Teixeira, F.Sprenger, S.P.Niclou, O.Keunen, L.V.M.Dixon, M.Williams, R.G.H.Beets-Tan, H.Franco-Maldonado, F.Loayza, J.Slotboom, P.Radojewski, R.Meier, R.Wiest, J.Trenkler, J.Pichler, G.Necker, S.Meckel, E.Torche, F.Vera, E.L\'{o}ópez, Y.Kim, H.Ismael, B.Allen, J.M.Buatti, J.Park, P.Zampakis, V.Panagiotopoulos, P.Tsiganos, E.Challiasos, D.M.Kardamakis, P.Prasanna, K.M.Mani, D.Payne, T.Kurc, L.Poisson, M.Valli\`{e}res, D.Fortin, M.Lepage, F.Mor\'{o}n, J.Mandel, C.Badve, A.E.Sloan, J.S.Barnholtz-Sloan, K.Waite, G.Shukla, S.Liem, G.S.Alexandre, J.Lombardo, J.D.Palmer, A.E.Flanders, A.P.Dicker, G.Ogbole, M.Soneye, D.Oyekunle, O.Odafe-Oyibotha, B.Osobu, M.Shu'aibu, F.Dako, A.Dorcas, D.Murcia, R.Haas, J.Thompson, D.R.Ormond, S.Currie, K.Fatania, R.Frood, J.Mitchell, J.Farinhas, A.L.Simpson, J.J.Peoples, R.Hu, D.Cutler, F.Y.Moraes, A.Tran, M.Hamghalam, M.A.Boss, J.Gimpel, B.Bialecki, A.Chelliah.
    
    \noindent \underline{\textbf{Data Processing:}} C.Sako, S.Ghodasara, E.Calabrese, J.Rudie, M.Jadhav, U.Pandey, R.Y.Huang, M.Jiang, C.Chen, C.Raymond, S.Bhardwaj, C.Chong, M.Agzarian, M.Kozubek, F.Lux, J.Mich\'{a}lek, P.Matula, C.Bangalore Yogananda, D.Reddy, B.C.Wagner, I.Ezhov, M.Lee, Y.W.Lui, R.Verma, R.Bareja, I.Yadav, J.Chen, N.Kumar, K.Bhuvaneshwar, A.Sayah, C.Bencheqroun, K.Kolodziej, M.Hill, M.Reyes, L.Pei, M.Ak, A.Kotrotsou, P.Vollmuth, G.Brugnara, C.J.Preetha, M.Zenk, J.Puig, M.Muzi, H.F.Shaykh, A.Abello, J.Bernal, J.G\'{o}mez, P.LaMontagne, K.Ramadass, S.Chotai, N.Wang, M.Smits, S.R.van der Voort, A.Alafandi, F.Incekara, M.M.J.Wijnenga, G.Kapsas, R.Gahrmann, A.J.P.E.Vincent, P.J.French, S.Klein, H.I.Sair, C.K.Jones, J.Garrett, H.Li, F.Kofler, Y.Yuan, S.Adabi, A.Xavier Falc\~{a}o, S.B.Martins, D.Menotti, D.R.Lucio, O.Keunen, A.Hau, K.Kamnitsas, L.Dixon, S.Benson, E.Pelaez, H.Franco-Maldonado, F.Loayza, S.Quevedo, R.McKinley, J.Trenkler, A.Haunschmidt, C.Mendoza, E.R\'{i}os, J.Choi, S.Baek, J.Yun, P.Zampakis, V.Panagiotopoulos, P.Tsiganos, E.I.Zacharaki, C.Kalogeropoulou, P.Prasanna, S.Shreshtra, T.Kurc, B.Luo, N.Wen, M.Valli\`{e}res, D.Fortin, F.Mor\'{o}n, C.Badve, V.Vadmal, G.Shukla, G.Ogbole, D.Oyekunle, F.Dako, D.Murcia, E.Fu, S.Currie, R.Frood, M.A.Vogelbaum, J.Mitchell, J.Farinhas, J.J.Peoples, M.Hamghalam, D.Kattil Veettil, K.Schmidt, B.Bialecki, S.Marella, T.C.Booth, A.Chelliah, M.Modat, C.Dragos, H.Shuaib.
    
    \noindent \underline{\textbf{Data Analysis \& Interpretation:}} S.Pati, U.Baid, B.Edwards, M.Sheller, S.Bakas. 
    
    \noindent \underline{\textbf{Site PI/Senior member (of each collaborating group):}} C.Davatzikos, J.Villanueva-Meyer, M.Ingalhalikar, R.Y.Huang, Q.Dou, B.M.Ellingson, M.To, M.Kozubek, J.A.Maldjian, B.Wiestler, R.Jain, P.Tiwari, Y.Gusev, A.Abayazeed, R.R.Colen, P.Vollmuth, A.Mahajan, C.Bala\~{n}a Quintero, S.Lee, M.Muzi, H.F.Shaykh, M.Trujillo, D.Marcus, B.Landman, A.Rao, M.Smits, H.I.Sair, R.Jeraj, B.Menze, Y.Yuan, A.Xavier Falc\~{a}o, S.P.Niclou, B.Glocker, J.Teuwen, E.Pelaez, R.Wiest, S.Meckel, P.Guevara, S.Baek, H.Kim, D.M.Kardamakis, J.Saltz, L.Poisson, M.Valli\`{e}res, F.Mor\'{o}n, A.E.Sloan, A.E.Flanders, G.Ogbole, D.R.Ormond, S.Currie, J.Farinhas, A.L.Simpson, C.Apgar, T.C.Booth.
    
    \noindent \underline{\textbf{Writing the Original Manuscript:}} S.Pati, U.Baid, B.Edwards, M.Sheller, S.Bakas.
    
    \noindent \underline{\textbf{Review, Edit, \& Approval of the Final Manuscript:}} All authors.


\section*{Competing Interest Declaration}
    The Intel affiliated authors (B.Edwards, M.Sheller, S.Wang, G.A.Reina, P.Foley, A.Gruzdev, D.Karkada, P.Shah, J.Martin) would like to disclose the following (potential) competing interest as Intel employees. Intel may develop proprietary software that is related in reputation to the OpenFL open source project highlighted in this work. In addition, the work demonstrates feasibility of federated learning for brain tumor boundary detection models. Intel may benefit by selling products to support an increase in demand for this use case.

           

\printbibliography

@article{brats_1,
  title={The multimodal brain tumor image segmentation benchmark (BRATS)},
  author={Menze, Bjoern H and Jakab, Andras and Bauer, Stefan and Kalpathy-Cramer, Jayashree and Farahani, Keyvan and Kirby, Justin and Burren, Yuliya and Porz, Nicole and Slotboom, Johannes and Wiest, Roland and others},
  journal={IEEE transactions on medical imaging},
  volume={34},
  number={10},
  pages={1993--2024},
  year={2014},
  publisher={IEEE}
}

@article{brats_2,
  title={Advancing the cancer genome atlas glioma MRI collections with expert segmentation labels and radiomic features},
  author={Bakas, Spyridon and Akbari, Hamed and Sotiras, Aristeidis and Bilello, Michel and Rozycki, Martin and Kirby, Justin S and Freymann, John B and Farahani, Keyvan and Davatzikos, Christos},
  journal={Scientific data},
  volume={4},
  number={1},
  pages={1--13},
  year={2017},
  publisher={Nature Publishing Group}
}

@article{brats_3,
  title={Identifying the best machine learning algorithms for brain tumor segmentation, progression assessment, and overall survival prediction in the BRATS challenge},
  author={Bakas, Spyridon and Reyes, Mauricio and Jakab, Andras and Bauer, Stefan and Rempfler, Markus and Crimi, Alessandro and Shinohara, Russell Takeshi and Berger, Christoph and Ha, Sung Min and Rozycki, Martin and others},
  journal={arXiv preprint arXiv:1811.02629},
  year={2018}
}

@article{brats_4,
  title={The rsna-asnr-miccai brats 2021 benchmark on brain tumor segmentation and radiogenomic classification},
  author={Baid, Ujjwal and Ghodasara, Satyam and Bilello, Michel and Mohan, Suyash and Calabrese, Evan and Colak, Errol and Farahani, Keyvan and Kalpathy-Cramer, Jayashree and Kitamura, Felipe C and Pati, Sarthak and others},
  journal={arXiv preprint arXiv:2107.02314},
  year={2021}
}

@article{louis20212021,
  title={The 2021 WHO classification of tumors of the central nervous system: a summary},
  author={Louis, David N and Perry, Arie and Wesseling, Pieter and Brat, Daniel J and Cree, Ian A and Figarella-Branger, Dominique and Hawkins, Cynthia and Ng, HK and Pfister, Stefan M and Reifenberger, Guido and others},
  journal={Neuro-oncology},
  volume={23},
  number={8},
  pages={1231--1251},
  year={2021},
  publisher={Oxford Academic}
}

@article{zech2018variable,
  title={Variable generalization performance of a deep learning model to detect pneumonia in chest radiographs: a cross-sectional study},
  author={Zech, John R and Badgeley, Marcus A and Liu, Manway and Costa, Anthony B and Titano, Joseph J and Oermann, Eric Karl},
  journal={PLoS medicine},
  volume={15},
  number={11},
  pages={e1002683},
  year={2018},
  publisher={Public Library of Science San Francisco, CA USA}
}

@article{maartensson2020reliability,
  title={The reliability of a deep learning model in clinical out-of-distribution MRI data: a multicohort study},
  author={M{\aa}rtensson, Gustav and Ferreira, Daniel and Granberg, Tobias and Cavallin, Lena and Oppedal, Ketil and Padovani, Alessandro and Rektorova, Irena and Bonanni, Laura and Pardini, Matteo and Kramberger, Milica G and others},
  journal={Medical Image Analysis},
  volume={66},
  pages={101714},
  year={2020},
  publisher={Elsevier}
}

@article{wilkinson2016fair,
  title={The FAIR Guiding Principles for scientific data management and stewardship},
  author={Wilkinson, Mark D and Dumontier, Michel and Aalbersberg, IJsbrand Jan and Appleton, Gabrielle and Axton, Myles and Baak, Arie and Blomberg, Niklas and Boiten, Jan-Willem and da Silva Santos, Luiz Bonino and Bourne, Philip E and others},
  journal={Scientific data},
  volume={3},
  number={1},
  pages={1--9},
  year={2016},
  publisher={Nature Publishing Group}
}

@article{kamnitsas2017efficient,
  title={Efficient multi-scale 3D CNN with fully connected CRF for accurate brain lesion segmentation},
  author={Kamnitsas, Konstantinos and Ledig, Christian and Newcombe, Virginia FJ and Simpson, Joanna P and Kane, Andrew D and Menon, David K and Rueckert, Daniel and Glocker, Ben},
  journal={Medical image analysis},
  volume={36},
  pages={61--78},
  year={2017},
  publisher={Elsevier}
}

@inproceedings{mckinley2018ensembles,
  title={Ensembles of densely-connected CNNs with label-uncertainty for brain tumor segmentation},
  author={McKinley, Richard and Meier, Raphael and Wiest, Roland},
  booktitle={International MICCAI Brainlesion Workshop},
  pages={456--465},
  year={2018},
  organization={Springer}
}

@article{isensee2021nnu,
  title={nnU-Net: a self-configuring method for deep learning-based biomedical image segmentation},
  author={Isensee, Fabian and Jaeger, Paul F and Kohl, Simon AA and Petersen, Jens and Maier-Hein, Klaus H},
  journal={Nature Methods},
  volume={18},
  number={2},
  pages={203--211},
  year={2021},
  publisher={Nature Publishing Group}
}

@article{dice,
  title={Morphometric analysis of white matter lesions in MR images: method and validation},
  author={Zijdenbos, Alex P and Dawant, Benoit M and Margolin, Richard A and Palmer, Andrew C},
  journal={IEEE transactions on medical imaging},
  volume={13},
  number={4},
  pages={716--724},
  year={1994},
  publisher={IEEE}
}

@inproceedings{horn1990hadamard,
  title={The hadamard product},
  author={Horn, Roger A},
  booktitle={Proc. Symp. Appl. Math},
  volume={40},
  pages={87--169},
  year={1990}
}

@article{barrodale1968l1,
  title={L1 approximation and the analysis of data},
  author={Barrodale, Ian},
  journal={Journal of the Royal Statistical Society: Series C (Applied Statistics)},
  volume={17},
  number={1},
  pages={51--57},
  year={1968},
  publisher={Wiley Online Library}
}

@article{ekberg2014untapped,
  title={The untapped potential of trusted execution environments on mobile devices},
  author={Ekberg, Jan-Erik and Kostiainen, Kari and Asokan, N},
  journal={IEEE Security \& Privacy},
  volume={12},
  number={4},
  pages={29--37},
  year={2014},
  publisher={IEEE}
}

@incollection{sudre2017generalised,
  title={Generalised dice overlap as a deep learning loss function for highly unbalanced segmentations},
  author={Sudre, Carole H and Li, Wenqi and Vercauteren, Tom and Ourselin, Sebastien and Cardoso, M Jorge},
  booktitle={Deep learning in medical image analysis and multimodal learning for clinical decision support},
  pages={240--248},
  year={2017},
  publisher={Springer}
}

@article{kingma2014adam,
  title={Adam: A method for stochastic optimization},
  author={Kingma, Diederik P and Ba, Jimmy},
  journal={arXiv preprint arXiv:1412.6980},
  year={2014}
}

@misc{labelfusion,
  author       = {Sarthak Pati and
                  Spyridon Bakas},
  title        = {{LabelFusion: Medical Image label fusion of 
                   segmentations}},
  month        = mar,
  year         = 2021,
  publisher    = {Zenodo},
  version      = {1.0.10}
}

@article{thakur2020brain,
  title={Brain extraction on MRI scans in presence of diffuse glioma: Multi-institutional performance evaluation of deep learning methods and robust modality-agnostic training},
  author={Thakur, Siddhesh and Doshi, Jimit and Pati, Sarthak and Rathore, Saima and Sako, Chiharu and Bilello, Michel and Ha, Sung Min and Shukla, Gaurav and Flanders, Adam and Kotrotsou, Aikaterini and others},
  journal={NeuroImage},
  volume={220},
  pages={117081},
  year={2020},
  publisher={Elsevier}
}

@article{pati2021gandlf,
  title={Gandlf: A generally nuanced deep learning framework for scalable end-to-end clinical workflows in medical imaging},
  author={Pati, Sarthak and Thakur, Siddhesh P and Bhalerao, Megh and Baid, Ujjwal and Grenko, Caleb and Edwards, Brandon and Sheller, Micah and Agraz, Jose and Baheti, Bhakti and Bashyam, Vishnu and others},
  journal={arXiv preprint arXiv:2103.01006},
  year={2021}
}

@inproceedings{salehi2017tversky,
  title={Tversky loss function for image segmentation using 3D fully convolutional deep networks},
  author={Salehi, Seyed Sadegh Mohseni and Erdogmus, Deniz and Gholipour, Ali},
  booktitle={International workshop on machine learning in medical imaging},
  pages={379--387},
  year={2017},
  organization={Springer}
}

@article{caliva2019distance,
  title={Distance map loss penalty term for semantic segmentation},
  author={Caliva, Francesco and Iriondo, Claudia and Martinez, Alejandro Morales and Majumdar, Sharmila and Pedoia, Valentina},
  journal={arXiv preprint arXiv:1908.03679},
  year={2019}
}

@article{chen2016efficient,
  title={Efficient and robust deep learning with correntropy-induced loss function},
  author={Chen, Liangjun and Qu, Hua and Zhao, Jihong and Chen, Badong and Principe, Jose C},
  journal={Neural Computing and Applications},
  volume={27},
  number={4},
  pages={1019--1031},
  year={2016},
  publisher={Springer}
}

@inproceedings{paszke2019pytorch,
  title={Pytorch: An imperative style, high-performance deep learning library},
  author={Paszke, Adam and Gross, Sam and Massa, Francisco and Lerer, Adam and Bradbury, James and Chanan, Gregory and Killeen, Trevor and Lin, Zeming and Gimelshein, Natalia and Antiga, Luca and others},
  booktitle={Advances in neural information processing systems},
  pages={8026--8037},
  year={2019}
}

@article{vakili2013enhanced,
  title={Enhanced precision analysis for accuracy-aware bit-width optimization using affine arithmetic},
  author={Vakili, Shervin and Langlois, JM Pierre and Bois, Guy},
  journal={IEEE Transactions on Computer-Aided Design of Integrated Circuits and Systems},
  volume={32},
  number={12},
  pages={1853--1865},
  year={2013},
  publisher={IEEE}
}

@inproceedings{lin2016fixed,
  title={Fixed point quantization of deep convolutional networks},
  author={Lin, Darryl and Talathi, Sachin and Annapureddy, Sreekanth},
  booktitle={International conference on machine learning},
  pages={2849--2858},
  year={2016},
  organization={PMLR}
}

@article{garcia_torchio_2020,
   title = {TorchIO: a Python library for efficient loading, preprocessing, augmentation and patch-based sampling of medical images in deep learning},
   journal = {Computer Methods and Programs in Biomedicine},
   pages = {106236},
   year = {2021},
   issn = {0169-2607},
   author = {P{\'e}rez-Garc{\'i}a, Fernando and Sparks, Rachel and Ourselin, S{\'e}bastien},
}

@article{young2020applicability,
  title={Applicability of various pre-trained deep convolutional neural networks for pneumonia classification based on X-Ray Images},
  author={Young, Julio Cristian and Suryadibrata, Alethea},
  journal={International Journal of Advanced Trends in Computer Science and Engineering},
  volume={9},
  number={3},
  year={2020},
  publisher={Warse}
}

@article{raghu2019transfusion,
  title={Transfusion: Understanding transfer learning for medical imaging},
  author={Raghu, Maithra and Zhang, Chiyuan and Kleinberg, Jon and Bengio, Samy},
  journal={Advances in neural information processing systems},
  volume={32},
  year={2019}
}

@article{yushkevich2016fast,
  title={Fast automatic segmentation of hippocampal subfields and medial temporal lobe subregions in 3 Tesla and 7 Tesla T2-weighted MRI},
  author={Yushkevich, Paul A and Pluta, John and Wang, Hongzhi and Wisse, Laura EM and Das, Sandhitsu and Wolk, David},
  journal={Alzheimer's \& Dementia: The Journal of the Alzheimer's Association},
  volume={12},
  number={7},
  pages={P126--P127},
  year={2016},
  publisher={Elsevier}
}

@article{py06nimg,
  author = {Paul A. Yushkevich and Joseph Piven and Cody Hazlett, Heather and
    Gimpel Smith, Rachel and Sean Ho and James C. Gee and Guido Gerig},
  title = {User-Guided {3D} Active Contour Segmentation of
    Anatomical Structures: Significantly Improved Efficiency and Reliability},
  journal = {Neuroimage},
  year = {2006},
  volume = {31},
  number = {3},
  pages = {1116--1128},
}

@article{captk_1,
  title={Cancer imaging phenomics toolkit: quantitative imaging analytics for precision diagnostics and predictive modeling of clinical outcome},
  author={Davatzikos, Christos and Rathore, Saima and Bakas, Spyridon and Pati, Sarthak and Bergman, Mark and Kalarot, Ratheesh and Sridharan, Patmaa and Gastounioti, Aimilia and Jahani, Nariman and Cohen, Eric and others},
  journal={Journal of medical imaging},
  volume={5},
  number={1},
  pages={011018},
  year={2018},
  publisher={International Society for Optics and Photonics}
}

@inproceedings{captk_2,
  title={Brain cancer imaging phenomics toolkit (brain-CaPTk): an interactive platform for quantitative analysis of glioblastoma},
  author={Rathore, Saima and Bakas, Spyridon and Pati, Sarthak and Akbari, Hamed and Kalarot, Ratheesh and Sridharan, Patmaa and Rozycki, Martin and Bergman, Mark and Tunc, Birkan and Verma, Ragini and others},
  booktitle={International MICCAI Brainlesion Workshop},
  pages={133--145},
  year={2017},
  organization={Springer}
}

@inproceedings{captk_3,
  title={The cancer imaging phenomics toolkit (captk): Technical overview},
  author={Pati, Sarthak and Singh, Ashish and Rathore, Saima and Gastounioti, Aimilia and Bergman, Mark and Ngo, Phuc and Ha, Sung Min and Bounias, Dimitrios and Minock, James and Murphy, Grayson and others},
  booktitle={International MICCAI Brainlesion Workshop},
  pages={380--394},
  year={2019},
  organization={Springer}
}

@inproceedings{bhalerao2019brain,
  title={Brain tumor segmentation based on 3D residual U-Net},
  author={Bhalerao, Megh and Thakur, Siddhesh},
  booktitle={International MICCAI Brainlesion Workshop},
  pages={218--225},
  year={2019},
  organization={Springer}
}

@inproceedings{ronneberger2015u,
  title={U-net: Convolutional networks for biomedical image segmentation},
  author={Ronneberger, Olaf and Fischer, Philipp and Brox, Thomas},
  booktitle={International Conference on Medical image computing and computer-assisted intervention},
  pages={234--241},
  year={2015},
  organization={Springer}
}

@inproceedings{cciccek20163d,
  title={3D U-Net: learning dense volumetric segmentation from sparse annotation},
  author={{\c{C}}i{\c{c}}ek, {\"O}zg{\"u}n and Abdulkadir, Ahmed and Lienkamp, Soeren S and Brox, Thomas and Ronneberger, Olaf},
  booktitle={International conference on medical image computing and computer-assisted intervention},
  pages={424--432},
  year={2016},
  organization={Springer}
}

@incollection{drozdzal2016importance,
  title={The importance of skip connections in biomedical image segmentation},
  author={Drozdzal, Michal and Vorontsov, Eugene and Chartrand, Gabriel and Kadoury, Samuel and Pal, Chris},
  booktitle={Deep Learning and Data Labeling for Medical Applications},
  pages={179--187},
  year={2016},
  publisher={Springer}
}

@inproceedings{he2016deep,
  title={Deep residual learning for image recognition},
  author={He, Kaiming and Zhang, Xiangyu and Ren, Shaoqing and Sun, Jian},
  booktitle={Proceedings of the IEEE conference on computer vision and pattern recognition},
  pages={770--778},
  year={2016}
}

@article{bakas2020iglass,
  title={iGLASS: imaging integration into the Glioma Longitudinal Analysis Consortium},
  author={Bakas, Spyridon and Ormond, David Ryan and Alfaro-Munoz, Kristin D and Smits, Marion and Cooper, Lee Alex Donald and Verhaak, Roel and Poisson, Laila M},
  journal={Neuro-oncology},
  volume={22},
  number={10},
  pages={1545--1546},
  year={2020},
  publisher={Oxford University Press US}
}

@article{glass2018glioma,
    author = {The GLASS Consortium },
    title = "{Glioma through the looking GLASS: molecular evolution of diffuse gliomas and the Glioma Longitudinal Analysis Consortium}",
    journal = {Neuro-Oncology},
    volume = {20},
    number = {7},
    pages = {873-884},
    year = {2018},
    month = {02},
    issn = {1522-8517}
}

@article{davatzikos2020ai,
  title={AI-based prognostic imaging biomarkers for precision neuro-oncology: the ReSPOND consortium},
  author={Davatzikos, Christos and Barnholtz-Sloan, Jill S and Bakas, Spyridon and Colen, Rivka and Mahajan, Abhishek and Quintero, Carmen Bala{\~n}a and Capellades Font, Jaume and Puig, Josep and Jain, Rajan and Sloan, Andrew E and others},
  journal={Neuro-oncology},
  volume={22},
  number={6},
  pages={886--888},
  year={2020},
  publisher={Oxford University Press US}
}

@article{thompson2014enigma,
  title={The ENIGMA Consortium: large-scale collaborative analyses of neuroimaging and genetic data},
  author={Thompson, Paul M and Stein, Jason L and Medland, Sarah E and Hibar, Derrek P and Vasquez, Alejandro Arias and Renteria, Miguel E and Toro, Roberto and Jahanshad, Neda and Schumann, Gunter and Franke, Barbara and others},
  journal={Brain imaging and behavior},
  volume={8},
  number={2},
  pages={153--182},
  year={2014},
  publisher={Springer}
}

@inproceedings{mcmahan2017communication,
  title={Communication-efficient learning of deep networks from decentralized data},
  author={McMahan, Brendan and Moore, Eider and Ramage, Daniel and Hampson, Seth and y Arcas, Blaise Aguera},
  booktitle={Artificial intelligence and statistics},
  pages={1273--1282},
  year={2017},
  organization={PMLR}
}

@article{shamir2019continuous,
  title={Continuous dice coefficient: a method for evaluating probabilistic segmentations},
  author={Shamir, Reuben R and Duchin, Yuval and Kim, Jinyoung and Sapiro, Guillermo and Harel, Noam},
  journal={arXiv preprint arXiv:1906.11031},
  year={2019}
}

@article{sheller2020federated,
  title={Federated learning in medicine: facilitating multi-institutional collaborations without sharing patient data},
  author={Sheller, Micah J and Edwards, Brandon and Reina, G Anthony and Martin, Jason and Pati, Sarthak and Kotrotsou, Aikaterini and Milchenko, Mikhail and Xu, Weilin and Marcus, Daniel and Colen, Rivka R and others},
  journal={Scientific reports},
  volume={10},
  number={1},
  pages={1--12},
  year={2020},
  publisher={Nature Publishing Group}
}

@article{rieke2020future,
  title={The future of digital health with federated learning},
  author={Rieke, Nicola and Hancox, Jonny and Li, Wenqi and Milletari, Fausto and Roth, Holger R and Albarqouni, Shadi and Bakas, Spyridon and Galtier, Mathieu N and Landman, Bennett A and Maier-Hein, Klaus and others},
  journal={NPJ digital medicine},
  volume={3},
  number={1},
  pages={1--7},
  year={2020},
  publisher={Nature Publishing Group}
}

@article{kaissis2020secure,
  title={Secure, privacy-preserving and federated machine learning in medical imaging},
  author={Kaissis, Georgios A and Makowski, Marcus R and R{\"u}ckert, Daniel and Braren, Rickmer F},
  journal={Nature Machine Intelligence},
  volume={2},
  number={6},
  pages={305--311},
  year={2020},
  publisher={Nature Publishing Group}
}

@article{dayan2021federated,
  title={Federated learning for predicting clinical outcomes in patients with COVID-19},
  author={Dayan, Ittai and Roth, Holger R and Zhong, Aoxiao and Harouni, Ahmed and Gentili, Amilcare and Abidin, Anas Z and Liu, Andrew and Costa, Anthony Beardsworth and Wood, Bradford J and Tsai, Chien-Sung and others},
  journal={Nature medicine},
  volume={27},
  number={10},
  pages={1735--1743},
  year={2021},
  publisher={Nature Publishing Group}
}

@incollection{roth2020federated,
  title={Federated learning for breast density classification: A real-world implementation},
  author={Roth, Holger R and Chang, Ken and Singh, Praveer and Neumark, Nir and Li, Wenqi and Gupta, Vikash and Gupta, Sharut and Qu, Liangqiong and Ihsani, Alvin and Bizzo, Bernardo C and others},
  booktitle={Domain Adaptation and Representation Transfer, and Distributed and Collaborative Learning},
  pages={181--191},
  year={2020},
  publisher={Springer}
}

@article{chang2018distributed,
  title={Distributed deep learning networks among institutions for medical imaging},
  author={Chang, Ken and Balachandar, Niranjan and Lam, Carson and Yi, Darvin and Brown, James and Beers, Andrew and Rosen, Bruce and Rubin, Daniel L and Kalpathy-Cramer, Jayashree},
  journal={Journal of the American Medical Informatics Association},
  volume={25},
  number={8},
  pages={945--954},
  year={2018},
  publisher={Oxford University Press}
}

@inproceedings{nilsson2018performance,
  title={A performance evaluation of federated learning algorithms},
  author={Nilsson, Adrian and Smith, Simon and Ulm, Gregor and Gustavsson, Emil and Jirstrand, Mats},
  booktitle={Proceedings of the Second Workshop on Distributed Infrastructures for Deep Learning},
  pages={1--8},
  year={2018}
}

@article{sarma2021federated,
  title={Federated learning improves site performance in multicenter deep learning without data sharing},
  author={Sarma, Karthik V and Harmon, Stephanie and Sanford, Thomas and Roth, Holger R and Xu, Ziyue and Tetreault, Jesse and Xu, Daguang and Flores, Mona G and Raman, Alex G and Kulkarni, Rushikesh and others},
  journal={Journal of the American Medical Informatics Association},
  volume={28},
  number={6},
  pages={1259--1264},
  year={2021},
  publisher={Oxford University Press}
}

@incollection{shen2021multi,
  title={Multi-task Federated Learning for Heterogeneous Pancreas Segmentation},
  author={Shen, Chen and Wang, Pochuan and Roth, Holger R and Yang, Dong and Xu, Daguang and Oda, Masahiro and Wang, Weichung and Fuh, Chiou-Shann and Chen, Po-Ting and Liu, Kao-Lang and others},
  booktitle={Clinical Image-Based Procedures, Distributed and Collaborative Learning, Artificial Intelligence for Combating COVID-19 and Secure and Privacy-Preserving Machine Learning},
  pages={101--110},
  year={2021},
  publisher={Springer}
}

@article{yang2021federated,
  title={Federated semi-supervised learning for COVID region segmentation in chest CT using multi-national data from China, Italy, Japan},
  author={Yang, Dong and Xu, Ziyue and Li, Wenqi and Myronenko, Andriy and Roth, Holger R and Harmon, Stephanie and Xu, Sheng and Turkbey, Baris and Turkbey, Evrim and Wang, Xiaosong and others},
  journal={Medical image analysis},
  volume={70},
  pages={101992},
  year={2021},
  publisher={Elsevier}
}

@article{ostrom2019cbtrus,
  title={CBTRUS statistical report: primary brain and other central nervous system tumors diagnosed in the United States in 2012--2016},
  author={Ostrom, Quinn T and Cioffi, Gino and Gittleman, Haley and Patil, Nirav and Waite, Kristin and Kruchko, Carol and Barnholtz-Sloan, Jill S},
  journal={Neuro-oncology},
  volume={21},
  number={Supplement\_5},
  pages={v1--v100},
  year={2019},
  publisher={Oxford University Press US}
}

@article{chaichana2013multi,
  title={Multi-institutional validation of a preoperative scoring system which predicts survival for patients with glioblastoma},
  author={Chaichana, Kaisorn L and Pendleton, Courtney and Chambless, Lola and Camara-Quintana, Joaquin and Nathan, Jay K and Hassam-Malani, Laila and Li, Gordon and Harsh IV, Griffith R and Thompson, Reid C and Lim, Michael and others},
  journal={Journal of Clinical Neuroscience},
  volume={20},
  number={10},
  pages={1422--1426},
  year={2013},
  publisher={Elsevier}
}

@article{fathi2020cancer,
  title={Cancer imaging phenomics via CaPTk: multi-institutional prediction of progression-free survival and pattern of recurrence in glioblastoma},
  author={Fathi Kazerooni, Anahita and Akbari, Hamed and Shukla, Gaurav and Badve, Chaitra and Rudie, Jeffrey D and Sako, Chiharu and Rathore, Saima and Bakas, Spyridon and Pati, Sarthak and Singh, Ashish and others},
  journal={JCO clinical cancer informatics},
  volume={4},
  pages={234--244},
  year={2020},
  publisher={American Society of Clinical Oncology}
}

@article{akbari2014pattern,
  title={Pattern analysis of dynamic susceptibility contrast-enhanced MR imaging demonstrates peritumoral tissue heterogeneity},
  author={Akbari, Hamed and Macyszyn, Luke and Da, Xiao and Wolf, Ronald L and Bilello, Michel and Verma, Ragini and O’Rourke, Donald M and Davatzikos, Christos},
  journal={Radiology},
  volume={273},
  number={2},
  pages={502--510},
  year={2014},
  publisher={Radiological Society of North America}
}

@article{akbari2020histopathology,
  title={Histopathology-validated machine learning radiographic biomarker for noninvasive discrimination between true progression and pseudo-progression in glioblastoma},
  author={Akbari, Hamed and Rathore, Saima and Bakas, Spyridon and Nasrallah, MacLean P and Shukla, Gaurav and Mamourian, Elizabeth and Rozycki, Martin and Bagley, Stephen J and Rudie, Jeffrey D and Flanders, Adam E and others},
  journal={Cancer},
  volume={126},
  number={11},
  pages={2625--2636},
  year={2020},
  publisher={Wiley Online Library}
}

@article{bakas2020overall,
  title={Overall survival prediction in glioblastoma patients using structural magnetic resonance imaging (MRI): advanced radiomic features may compensate for lack of advanced MRI modalities},
  author={Bakas, Spyridon and Shukla, Gaurav and Akbari, Hamed and Erus, Guray and Sotiras, Aristeidis and Rathore, Saima and Sako, Chiharu and Ha, Sung Min and Rozycki, Martin and Shinohara, Russell T and others},
  journal={Journal of Medical Imaging},
  volume={7},
  number={3},
  pages={031505},
  year={2020},
  publisher={International Society for Optics and Photonics}
}

@article{binder2018epidermal,
  title={Epidermal growth factor receptor extracellular domain mutations in glioblastoma present opportunities for clinical imaging and therapeutic development},
  author={Binder, Zev A and Thorne, Amy Haseley and Bakas, Spyridon and Wileyto, E Paul and Bilello, Michel and Akbari, Hamed and Rathore, Saima and Ha, Sung Min and Zhang, Logan and Ferguson, Cole J and others},
  journal={Cancer cell},
  volume={34},
  number={1},
  pages={163--177},
  year={2018},
  publisher={Elsevier}
}

@article{akbari2018vivo,
  title={In vivo evaluation of EGFRvIII mutation in primary glioblastoma patients via complex multiparametric MRI signature},
  author={Akbari, Hamed and Bakas, Spyridon and Pisapia, Jared M and Nasrallah, MacLean P and Rozycki, Martin and Martinez-Lage, Maria and Morrissette, Jennifer JD and Dahmane, Nadia and O’Rourke, Donald M and Davatzikos, Christos},
  journal={Neuro-oncology},
  volume={20},
  number={8},
  pages={1068--1079},
  year={2018},
  publisher={Oxford University Press US}
}

@article{mang2020integrated,
  title={Integrated biophysical modeling and image analysis: application to neuro-oncology},
  author={Mang, Andreas and Bakas, Spyridon and Subramanian, Shashank and Davatzikos, Christos and Biros, George},
  journal={Annual review of biomedical engineering},
  volume={22},
  pages={309--341},
  year={2020},
  publisher={Annual Reviews}
}

@article{akbari2016imaging,
  title={Imaging surrogates of infiltration obtained via multiparametric imaging pattern analysis predict subsequent location of recurrence of glioblastoma},
  author={Akbari, Hamed and Macyszyn, Luke and Da, Xiao and Bilello, Michel and Wolf, Ronald L and Martinez-Lage, Maria and Biros, George and Alonso-Basanta, Michelle and O'Rourke, Donald M and Davatzikos, Christos},
  journal={Neurosurgery},
  volume={78},
  number={4},
  pages={572--580},
  year={2016},
  publisher={Oxford University Press}
}

@article{shukla2017advanced,
  title={Advanced magnetic resonance imaging in glioblastoma: a review},
  author={Shukla, Gaurav and Alexander, Gregory S and Bakas, Spyridon and Nikam, Rahul and Talekar, Kiran and Palmer, Joshua D and Shi, Wenyin},
  journal={Chin Clin Oncol},
  volume={6},
  number={4},
  pages={40},
  year={2017}
}

@article{aggarwal2018neural,
  title={Neural networks and deep learning},
  author={Aggarwal, Charu C and others},
  journal={Springer},
  volume={10},
  pages={978--3},
  year={2018},
  publisher={Springer}
}

@article{marcus2018deep,
  title={Deep learning: A critical appraisal},
  author={Marcus, Gary},
  journal={arXiv preprint arXiv:1801.00631},
  year={2018}
}

@article{obermeyer2016predicting,
  title={Predicting the future—big data, machine learning, and clinical medicine},
  author={Obermeyer, Ziad and Emanuel, Ezekiel J},
  journal={The New England journal of medicine},
  volume={375},
  number={13},
  pages={1216},
  year={2016},
  publisher={NIH Public Access}
}

@article{annas2003hipaa,
  title={HIPAA regulations-a new era of medical-record privacy?},
  author={Annas, George J and others},
  journal={New England Journal of Medicine},
  volume={348},
  number={15},
  pages={1486--1490},
  year={2003},
  publisher={MEDICAL PUBLISHING GROUP-MASS MEDIC SOCIETY}
}

@article{voigt2017eu,
  title={The eu general data protection regulation (gdpr)},
  author={Voigt, Paul and Von dem Bussche, Axel},
  journal={A Practical Guide, 1st Ed., Cham: Springer International Publishing},
  volume={10},
  pages={3152676},
  year={2017},
  publisher={Springer}
}

@article{reina2021openfl,
  title={OpenFL: An open-source framework for Federated Learning},
  author={Reina, G Anthony and Gruzdev, Alexey and Foley, Patrick and Perepelkina, Olga and Sharma, Mansi and Davidyuk, Igor and Trushkin, Ilya and Radionov, Maksim and Mokrov, Aleksandr and Agapov, Dmitry and others},
  journal={arXiv preprint arXiv:2105.06413},
  year={2021}
}

@inproceedings{gorbachev2019openvino,
  title={OpenVINO deep learning workbench: Comprehensive analysis and tuning of neural networks inference},
  author={Gorbachev, Yury and Fedorov, Mikhail and Slavutin, Iliya and Tugarev, Artyom and Fatekhov, Marat and Tarkan, Yaroslav},
  booktitle={Proceedings of the IEEE/CVF International Conference on Computer Vision Workshops},
  pages={783--787},
  year={2019}
}

@article{rodriguez2018lower,
  title={Lower numerical precision deep learning inference and training},
  author={Rodriguez, Andres and Segal, Eden and Meiri, Etay and Fomenko, Evarist and Kim, Y Jim and Shen, Haihao and Ziv, Barukh},
  journal={Intel White Paper},
  volume={3},
  pages={1--19},
  year={2018}
}

@article{bakas2017vivo,
  title={In vivo detection of EGFRvIII in glioblastoma via perfusion magnetic resonance imaging signature consistent with deep peritumoral infiltration: the $\varphi$-index},
  author={Bakas, Spyridon and Akbari, Hamed and Pisapia, Jared and Martinez-Lage, Maria and Rozycki, Martin and Rathore, Saima and Dahmane, Nadia and O'Rourke, Donald M and Davatzikos, Christos},
  journal={Clinical Cancer Research},
  volume={23},
  number={16},
  pages={4724--4734},
  year={2017},
  publisher={AACR}
}

@article{fathi2020imaging,
  title={Imaging signatures of glioblastoma molecular characteristics: a radiogenomics review},
  author={Fathi Kazerooni, Anahita and Bakas, Spyridon and Saligheh Rad, Hamidreza and Davatzikos, Christos},
  journal={Journal of Magnetic Resonance Imaging},
  volume={52},
  number={1},
  pages={54--69},
  year={2020},
  publisher={Wiley Online Library}
}

@article{bakas2018nimg,
  title={NIMG-40. Non-invasive in vivo signature of idh1 mutational status in high grade glioma, from clinically-acquired multi-parametric magnetic resonance imaging, using multivariate machine learning},
  author={Bakas, Spyridon and Rathore, Saima and Nasrallah, MacLean and Akbari, Hamed and Binder, Zev and Ha, Sung Min and Mamourian, Elizabeth and Morrissette, Jennifer and O’Rourke, Donald and Davatzikos, Christos},
  journal={Neuro-Oncology},
  volume={20},
  number={suppl\_6},
  pages={vi184--vi185},
  year={2018},
  publisher={Oxford University Press US}
}

@inproceedings{sheller2018multiinstitutional,
  title={Multi-institutional deep learning modeling without sharing patient data: A feasibility study on brain tumor segmentation},
  author={Sheller, Micah J and Reina, G Anthony and Edwards, Brandon and Martin, Jason and Bakas, Spyridon},
  booktitle={International MICCAI Brainlesion Workshop},
  pages={92--104},
  year={2018},
  organization={Springer}
}

@inproceedings{rathore2018deriving,
  title={Deriving stable multi-parametric MRI radiomic signatures in the presence of inter-scanner variations: survival prediction of glioblastoma via imaging pattern analysis and machine learning techniques},
  author={Rathore, Saima and Bakas, Spyridon and Akbari, Hamed and Shukla, Gaurav and Rozycki, Martin and Davatzikos, Christos},
  booktitle={Medical Imaging 2018: Computer-Aided Diagnosis},
  volume={10575},
  pages={1057509},
  year={2018},
  organization={International Society for Optics and Photonics}
}

@article{brennan2013somatic,
  title={The somatic genomic landscape of glioblastoma},
  author={Brennan, Cameron W and Verhaak, Roel GW and McKenna, Aaron and Campos, Benito and Noushmehr, Houtan and Salama, Sofie R and Zheng, Siyuan and Chakravarty, Debyani and Sanborn, J Zachary and Berman, Samuel H and others},
  journal={Cell},
  volume={155},
  number={2},
  pages={462--477},
  year={2013},
  publisher={Elsevier}
}

@article{verhaak2010integrated,
  title={Integrated genomic analysis identifies clinically relevant subtypes of glioblastoma characterized by abnormalities in PDGFRA, IDH1, EGFR, and NF1},
  author={Verhaak, Roel GW and Hoadley, Katherine A and Purdom, Elizabeth and Wang, Victoria and Qi, Yuan and Wilkerson, Matthew D and Miller, C Ryan and Ding, Li and Golub, Todd and Mesirov, Jill P and others},
  journal={Cancer cell},
  volume={17},
  number={1},
  pages={98--110},
  year={2010},
  publisher={Elsevier}
}

@article{sottoriva2013intratumor,
  title={Intratumor heterogeneity in human glioblastoma reflects cancer evolutionary dynamics},
  author={Sottoriva, Andrea and Spiteri, Inmaculada and Piccirillo, Sara GM and Touloumis, Anestis and Collins, V Peter and Marioni, John C and Curtis, Christina and Watts, Colin and Tavar{\'e}, Simon},
  journal={Proceedings of the National Academy of Sciences},
  volume={110},
  number={10},
  pages={4009--4014},
  year={2013},
  publisher={National Acad Sciences}
}

@article{han2020deep,
  title={Deep transfer learning and radiomics feature prediction of survival of patients with high-grade gliomas},
  author={Han, W and Qin, L and Bay, C and Chen, X and Yu, K-H and Miskin, N and Li, A and Xu, X and Young, G},
  journal={American Journal of Neuroradiology},
  volume={41},
  number={1},
  pages={40--48},
  year={2020},
  publisher={Am Soc Neuroradiology}
}

@article{griggs2009rareDisease,
  title={Clinical research for rare disease: opportunities, challenges, and solutions},
  author={Griggs, Robert C and Batshaw, Mark and Dunkle, Mary and Gopal-Srivastava, Rashmi and Kaye, Edward and Krischer, Jeffrey and Nguyen, Tan and Paulus, Kathleen and Merkel, Peter A and others},
  journal={Molecular genetics and metabolism},
  volume={96},
  number={1},
  pages={20--26},
  year={2009},
  publisher={Elsevier}
}

@article{rudie2019multi,
  title={Multi-disease segmentation of gliomas and white matter hyperintensities in the BraTS data using a 3D convolutional neural network},
  author={Rudie, Jeffrey D and Weiss, David A and Saluja, Rachit and Rauschecker, Andreas M and Wang, Jiancong and Sugrue, Leo and Bakas, Spyridon and Colby, John B},
  journal={Frontiers in Computational Neuroscience},
  volume={13},
  pages={84},
  year={2019},
  publisher={Frontiers}
}

@article{natMed2018Retina,
  title={Clinically applicable deep learning for diagnosis and referral in retinal disease},
  author={De Fauw, Jeffrey and Ledsam, Joseph R and Romera-Paredes, Bernardino and Nikolov, Stanislav and Tomasev, Nenad and Blackwell, Sam and Askham, Harry and Glorot, Xavier and O’Donoghue, Brendan and Visentin, Daniel and others},
  journal={Nature medicine},
  volume={24},
  number={9},
  pages={1342--1350},
  year={2018},
  publisher={Nature Publishing Group}
}

@article{natMed2019Cardiac,
  title={Cardiologist-level arrhythmia detection and classification in ambulatory electrocardiograms using a deep neural network},
  author={Hannun, Awni Y and Rajpurkar, Pranav and Haghpanahi, Masoumeh and Tison, Geoffrey H and Bourn, Codie and Turakhia, Mintu P and Ng, Andrew Y},
  journal={Nature medicine},
  volume={25},
  number={1},
  pages={65--69},
  year={2019},
  publisher={Nature Publishing Group}
}

@article{RTOG0825_2014_NEJM,
  title={A randomized trial of bevacizumab for newly diagnosed glioblastoma},
  author={Gilbert, Mark R and Dignam, James J and Armstrong, Terri S and Wefel, Jeffrey S and Blumenthal, Deborah T and Vogelbaum, Michael A and Colman, Howard and Chakravarti, Arnab and Pugh, Stephanie and Won, Minhee and others},
  journal={New England Journal of Medicine},
  volume={370},
  number={8},
  pages={699--708},
  year={2014},
  publisher={Mass Medical Soc}
}

@misc{RTOG_2013_abstract,
  title={RTOG 0825: Phase III double-blind placebo-controlled trial evaluating bevacizumab (Bev) in patients (Pts) with newly diagnosed glioblastoma (GBM).},
  author={Gilbert, Mark R and Dignam, James and Won, Minhee and Blumenthal, Deborah T and Vogelbaum, Michael A and Aldape, Kenneth D and Colman, Howard and Chakravarti, Arnab and Jeraj, Robert and Armstrong, Terri S and others},
  year={2013},
  publisher={American Society of Clinical Oncology}
}

@article{ACRIN6686_2021_NOnc,
  title={Value of dynamic contrast perfusion MRI to predict early response to bevacizumab in newly diagnosed glioblastoma: results from ACRIN 6686 multicenter trial},
  author={Schmainda, Kathleen M and Prah, Melissa A and Marques, Helga and Kim, Eunhee and Barboriak, Daniel P and Boxerman, Jerrold L},
  journal={Neuro-oncology},
  volume={23},
  number={2},
  pages={314--323},
  year={2021},
  publisher={Oxford University Press US}
}

@article{ACRIN6686_2018_NOnc,
  title={Prognostic value of contrast enhancement and FLAIR for survival in newly diagnosed glioblastoma treated with and without bevacizumab: results from ACRIN 6686},
  author={Boxerman, Jerrold L and Zhang, Zheng and Safriel, Yair and Rogg, Jeffrey M and Wolf, Ronald L and Mohan, Suyash and Marques, Helga and Sorensen, A Gregory and Gilbert, Mark R and Barboriak, Daniel P},
  journal={Neuro-oncology},
  volume={20},
  number={10},
  pages={1400--1410},
  year={2018},
  publisher={Oxford University Press US}
}

@inproceedings{bakas2015glistrboost,
  title={GLISTRboost: combining multimodal MRI segmentation, registration, and biophysical tumor growth modeling with gradient boosting machines for glioma segmentation},
  author={Bakas, Spyridon and Zeng, Ke and Sotiras, Aristeidis and Rathore, Saima and Akbari, Hamed and Gaonkar, Bilwaj and Rozycki, Martin and Pati, Sarthak and Davatzikos, Christos},
  booktitle={BrainLes 2015},
  pages={144--155},
  year={2015},
  organization={Springer}
}

@inproceedings{zeng2016segmentation,
  title={Segmentation of gliomas in pre-operative and post-operative multimodal magnetic resonance imaging volumes based on a hybrid generative-discriminative framework},
  author={Zeng, Ke and Bakas, Spyridon and Sotiras, Aristeidis and Akbari, Hamed and Rozycki, Martin and Rathore, Saima and Pati, Sarthak and Davatzikos, Christos},
  booktitle={International Workshop on Brainlesion: Glioma, Multiple Sclerosis, Stroke and Traumatic Brain Injuries},
  pages={184--194},
  year={2016},
  organization={Springer}
}

@article{pei2020longitudinal,
  title={Longitudinal brain tumor segmentation prediction in MRI using feature and label fusion},
  author={Pei, Linmin and Bakas, Spyridon and Vossough, Arastoo and Reza, Syed MS and Davatzikos, Christos and Iftekharuddin, Khan M},
  journal={Biomedical signal processing and control},
  volume={55},
  pages={101648},
  year={2020},
  publisher={Elsevier}
}

@article{bakas2015segmentation,
  title={Segmentation of gliomas in multimodal magnetic resonance imaging volumes based on a hybrid generative-discriminative framework},
  author={Bakas, Spyridon and Zeng, Ke and Sotiras, Aristeidis and Rathore, Saima and Akbari, Hamed and Gaonkar, Bilwaj and Rozycki, Martin and Pati, Sarthak and Davazikos, C},
  journal={Proceeding of the Multimodal Brain Tumor Image Segmentation Challenge},
  pages={5--12},
  year={2015}
}

@article{menze2021analyzing,
  title={Analyzing magnetic resonance imaging data from glioma patients using deep learning},
  author={Menze, Bjoern and Isensee, Fabian and Wiest, Roland and Wiestler, Bene and Maier-Hein, Klaus and Reyes, Mauricio and Bakas, Spyridon},
  journal={Computerized medical imaging and graphics},
  volume={88},
  pages={101828},
  year={2021},
  publisher={Elsevier}
}

@article{pisapia2018use,
  title={Use of fetal magnetic resonance image analysis and machine learning to predict the need for postnatal cerebrospinal fluid diversion in fetal ventriculomegaly},
  author={Pisapia, Jared M and Akbari, Hamed and Rozycki, Martin and Goldstein, Hannah and Bakas, Spyridon and Rathore, Saima and Moldenhauer, Julie S and Storm, Phillip B and Zarnow, Deborah M and Anderson, Richard CE and others},
  journal={JAMA pediatrics},
  volume={172},
  number={2},
  pages={128--135},
  year={2018},
  publisher={American Medical Association}
}

@article{davatzikos2019precision,
  title={Precision diagnostics based on machine learning-derived imaging signatures},
  author={Davatzikos, Christos and Sotiras, Aristeidis and Fan, Yong and Habes, Mohamad and Erus, Guray and Rathore, Saima and Bakas, Spyridon and Chitalia, Rhea and Gastounioti, Aimilia and Kontos, Despina},
  journal={Magnetic resonance imaging},
  volume={64},
  pages={49--61},
  year={2019},
  publisher={Elsevier}
}

@article{han2018brain,
  title={Brain extraction from normal and pathological images: A joint PCA/image-reconstruction approach},
  author={Han, Xu and Kwitt, Roland and Aylward, Stephen and Bakas, Spyridon and Menze, Bjoern and Asturias, Alexander and Vespa, Paul and Van Horn, John and Niethammer, Marc},
  journal={NeuroImage},
  volume={176},
  pages={431--445},
  year={2018},
  publisher={Elsevier}
}

\end{document}